%% file: main.tex
\documentclass[11pt,letterpaper]{article}
\usepackage[margin=1in]{geometry}
\usepackage[skip=4pt,indent=0pt]{parskip}
\usepackage[T1]{fontenc}
\usepackage[utf8]{inputenc}
\usepackage{lmodern}
\usepackage[authoryear,round]{natbib}
\setcitestyle{authoryear,round,citesep={;},aysep={,},yysep={;}}
\renewcommand{\bibfont}{\small\raggedright}
\input{math_commands.tex}

\usepackage{amsmath,amssymb}
\usepackage{color,colortbl}
\usepackage{graphicx}
\usepackage{adjustbox}
\usepackage{booktabs}
\usepackage{tabularx}
\usepackage{array}
\usepackage{listings}
\usepackage{flafter}
\usepackage{float}
\usepackage{caption}
\usepackage{placeins}
\usepackage{microtype}
\usepackage{xurl}
\usepackage[most]{tcolorbox}
\usepackage{enumitem}
\usepackage{capt-of}
\usepackage{multirow}
\usepackage{makecell}
\newtcolorbox{findingbox}{
  colback=black!3,
  colframe=black!70,
  boxrule=0.8pt,
  arc=2mm,
  left=3mm,
  right=3mm,
  top=2mm,
  bottom=2mm,
  before skip=8pt,
  after skip=8pt
}
\tcbuselibrary{skins}
\newtcolorbox{textpromptbox}{
  enhanced,
  frame hidden,
  interior hidden,
  borderline west={1.2pt}{0pt}{black!25},
  sharp corners,
  boxrule=0pt,
  boxsep=0pt,
  left=3mm,
  right=1mm,
  top=1mm,
  bottom=1mm,
  before skip=6pt,
  after skip=6pt,
  fontupper=\small,
  colupper=black!80
}

\usepackage{tikz}
\usetikzlibrary{arrows.meta, positioning, fit}
\usepackage{hyperref}
\hypersetup{hidelinks}
\hypersetup{pdftitle={Grounding Generative Verification in 3D Content}, pdfauthor={Shuanglin Li, Jiaying Lin, Shuquan Ye}}

\definecolor{humanheader}{RGB}{232,232,248}
\definecolor{baselineheader}{RGB}{245,245,220}
\definecolor{closedheader}{RGB}{250,244,214}
\definecolor{openheader}{RGB}{224,246,246}

\definecolor{oursBlueFill}{RGB}{238,247,255}
\definecolor{oursBlueHeader}{RGB}{220,238,253}
\definecolor{oursBlueBorder}{RGB}{73,148,214}
\definecolor{oursBlueText}{RGB}{24,82,145}
\definecolor{oursBlueArrow}{RGB}{69,145,210}

\newcolumntype{Y}{>{\raggedleft\arraybackslash}X}
\title{Grounding Generative Verification in 3D Content}
\author{
  Shuanglin Li\textsuperscript{1,2}\quad
  Jiaying Lin\textsuperscript{3}\quad
  Shuquan Ye\textsuperscript{2}\\[0.5em]
  \small \textsuperscript{1}Southeast University\\
  \small \textsuperscript{2}Shenzhen Loop Area Institute\\
  \small \textsuperscript{3}The Hong Kong University of Science and Technology
}
\date{}

\begin{document}
\maketitle

\begin{abstract}
\noindent
Recent advances in multimodal large language models (MLLMs) and 3D generation have significantly expanded AI capabilities in understanding and synthesizing 3D content. However, beyond visual fidelity, task-specific correctness grounded in 3D evidence remains a major challenge. We argue that building reliable 3D systems requires not only stronger generation and reasoning, but also verification feedback that identifies unmet requirements and explains their violations, allowing systems to reflect on and refine their outputs. Such feedback must integrate multi-view observations and distinguish genuine inconsistencies from projection-dependent appearances. In this work, we make three contributions. First, we introduce 3DVerifierBench, a benchmark spanning 10 tasks that evaluates decisions, rejection explanations, and responses to controlled changes in specifications and candidates. A human study reveals a 21.6 percentage-point gap between humans and the strongest tested model, GPT-6 Astra, while explanation-aware analysis exposes weaknesses in rejection reliability. Second, we analyze verification across fine-grained factual categories, exposing errors in values and target binding. We develop a scalable pipeline combining multi-view fact extraction, controlled 3D editing, and native annotations to produce 14,126 training examples with contradiction explanations and linked counterfactuals. Third, we examine how verification could support 3D generation through candidate selection and recursive refinement. In human-labeled sixteen-candidate pools, oracle selection improves complete factual success from 26.7\% to 90.0\%. In a separate eight-input study, human-guided recursive refinement satisfies all seven requirements for every input, compared with three under parallel Oracle@8. Together, our benchmark, analysis, and data pipeline establish a foundation for models to reflect on 3D content through reliable verification feedback.

\end{abstract}

\vspace{-0.5em}
\begin{quotation}
\noindent\small
\textbf{Project page:}
\href{https://ai-yeslab.github.io/3DVerifier-page/}
{\textcolor{blue}{\nolinkurl{https://ai-yeslab.github.io/3DVerifier-page/}}}
\end{quotation}
\vspace{-1em}

\begin{figure}[ht]
\centering\includegraphics[width=\textwidth]{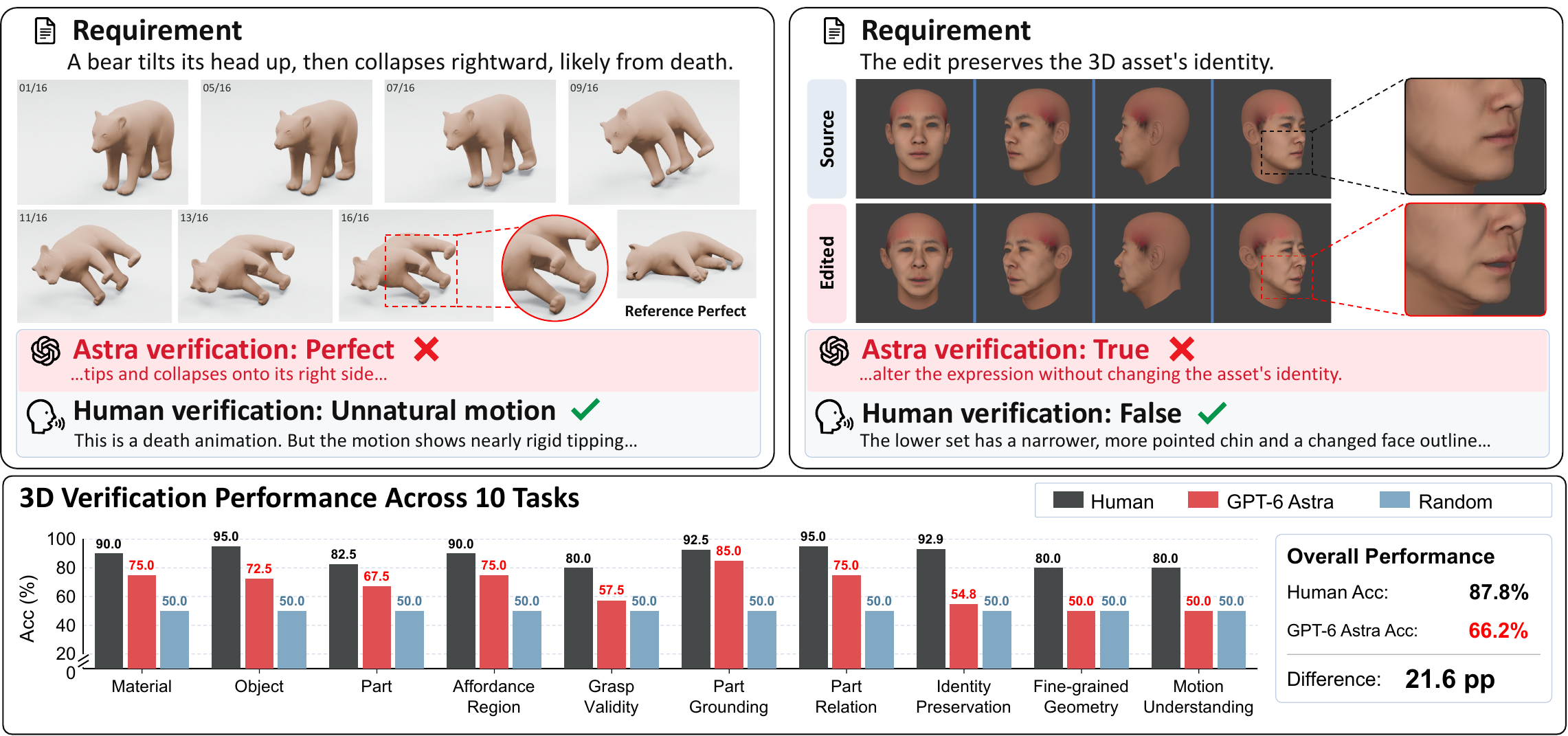}
\caption{\textbf{Even state-of-the-art MLLMs, \textit{e.g.}, GPT-6 Astra, still struggle with 3D content verification. }
Top: Astra mistakenly accepts an unnatural rightward collapse and an identity-changing face edit. 
Bottom: We found that Astra is still far behind human performance across all tasks in our 3DVerifierBench.
Views and text are abbreviated for display.}
\label{fig:teaser}
\end{figure}

\section{Introduction}\label{introduction}

Recent advances in multimodal large language models (MLLMs) have extended visual understanding toward increasingly complex spatial reasoning, while progress in 3D generation enables the synthesis and editing of assets with detailed geometry and appearance \citep{xu2024pointllm,deng2025_3dllava,li2026pixal3d}. Together, these developments bring models closer to understanding, creating, and manipulating 3D content through natural language. However, generated assets can still contain geometric artifacts, missing components, and inconsistencies with the requested attributes or structure \citep{wu2024gpteval3d,duggal2025eval3d}. Such errors limit their usefulness: An object may look convincing while failing the geometric or functional requirement for which it was created.

To improve these systems, existing research has pursued broader pretraining, targeted post-training, and test-time scaling. These approaches strengthen perceptual representations, align outputs with instructions or preferences, and allocate additional computation to candidate search \citep{xu2024pointllm,deng2025_3dllava,ye2024dreamreward,zhou2025its3d}. A system can also reflect on its output and use feedback to guide a subsequent attempt \citep{madaan2023selfrefine,shinn2023reflexion}. For 3D content, this requires identifying which requirements are unmet and explaining why. We study this step through \textbf{3D verification}, which checks a candidate against explicit requirements and explains contradictions. Despite progress in general visual verification \citep{zhang2025omniverifier}, the reliability of such feedback across diverse 3D tasks remains underexplored.

Realizing this goal in 3D, however, involves challenges beyond those of image-based verification. Native 3D perception requires specialized representations and language alignment, as illustrated by point-cloud language models \citep{xu2024pointllm,deng2025_3dllava}; these capabilities cannot simply be assumed in an image-based MLLM. Although multi-view rendering makes 3D assets accessible to such models, it provides partial, projection-dependent observations rather than the complete geometry. A verifier must consequently associate parts across views, resolve occlusion, and distinguish geometric differences from changes in viewpoint or appearance. For grasping, identity, and motion, it must further establish whether the specific constraint holds, rather than merely recognize the object or action. The two cases in Figure~\ref{fig:teaser} illustrate this distinction: GPT-6 Astra recognizes a rightward collapse but misses its unnatural motion, and accepts identity preservation despite a changed chin and face outline. These failures motivate both a systematic assessment of current verification ability and supervision that connects judgments to the relevant 3D evidence.

Motivated by these challenges, our work addresses three connected questions, from measuring the capability gap to supporting its development and examining its downstream value.

\textbf{How far are current MLLMs from human-level 3D verification?} We introduce \textbf{3DVerifierBench}, a human-reviewed benchmark with 2,581 examples across ten complementary tasks. A common answer-and-explanation interface covers object properties, parts, relations, affordance regions, grasps, identity, reference geometry, and motion. In addition to individual judgments, linked counterfactuals test whether predictions respond consistently to changes in requirements or candidates. Our experiments show that humans achieve 87.8\% task overall accuracy, exceeding the strongest tested model, GPT-6 Astra, by \textbf{21.6 percentage points} (Figure~\ref{fig:teaser}, bottom). The benchmark thus establishes both the breadth and the remaining limitations of current 3D verification.

\textbf{What capabilities and supervision are needed to strengthen 3D verification?} Our analysis shows that recognizing an object or action does not ensure verification of the decisive attribute, target, or geometric condition. Identity preservation, grasp validity, and motion compliance are particularly weak, while explanation checks reveal that even correct rejections can miss the reference contradiction. These findings motivate supervision that links a decision to a specific fact and its evidence. Motivated also by prior work on training multimodal verifiers~\citep{zhang2025omniverifier}, we develop an automated pipeline for large-scale verification data curation, combining multi-view fact extraction, controlled requirement and candidate changes, and native task annotations. With this pipeline, we produce a large-scale dataset with \textbf{14,126 samples} across diverse tasks, which provides a data foundation for scaling domain-specific verification supervision.

\textbf{How can a 3D generation model reflect by verifying reliably?} We examine both selection among existing candidates and recursive refinement of unsatisfied requirements. In human-labeled sixteen-candidate pools, oracle selection improves complete factual success by \textbf{63.3 percentage points}. In a separate eight-input study, human-guided recursive refinement satisfies all seven requirements for every input, compared with three under parallel Oracle@8. Together, these results demonstrate substantial downstream headroom for reliable verification, connecting the capability gap and data foundation to the goal of reflection in 3D generation.

\section{Related Work}\label{related-work}

\begin{figure}[!htbp]
\captionsetup{font=small}
\centering
\begin{tikzpicture}[
    font=\small,
    node distance=3.2mm and 4.0mm,
    prior/.style={rounded corners=3pt,draw=black!25,line width=0.6pt,fill=black!3,align=left,text width=0.205\linewidth,inner xsep=6.5pt,inner ysep=6pt,minimum height=14mm},
    gap/.style={rounded corners=3pt,draw=orange!55!black,line width=0.65pt,fill=orange!8,align=left,text width=0.300\linewidth,inner xsep=7pt,inner ysep=6pt,minimum height=14mm},
    ours/.style={rounded corners=3pt,draw=oursBlueBorder,line width=0.7pt,fill=oursBlueFill,align=left,text width=0.330\linewidth,inner xsep=7pt,inner ysep=6pt,minimum height=14mm},
    header/.style={font=\small\bfseries,text=black,align=center,inner sep=0pt},
    arrowleft/.style={-{Latex[length=2.0mm,width=1.5mm]},draw=black!38,line width=0.7pt},
    arrowours/.style={-{Latex[length=2.0mm,width=1.5mm]},draw=oursBlueArrow,line width=0.8pt}
]
\node[prior] (p1) {\textbf{3DGen-Bench}\\[-1pt] Human preference};
\node[gap, right=4.0mm of p1] (g1) {Satisfaction of individual requirements};
\node[ours, right=4.0mm of g1] (o1) {Explicit claims with binary correctness labels};
\node[header, above=0.6mm of p1] (h1) {Prior evaluation};
\node[header, above=0.6mm of g1] (h2) {Not directly assessed};
\node[header, above=0.6mm of o1] (h3) {3DVerifierBench};
\draw[arrowleft] (p1.east) -- (g1.west);
\draw[arrowours] (g1.east) -- (o1.west);
\node[prior, below=1mm of p1] (p2) {\textbf{Hi3DEval}\\[-1pt] Hierarchical quality};
\node[gap, right=4.0mm of p2] (g2) {Whether judgments track controlled changes to the requirement or candidate};
\node[ours, right=4.0mm of g2] (o2) {Linked Spec-CF / Cand-CF pairs and triplet success};
\draw[arrowleft] (p2.east) -- (g2.west);
\draw[arrowours] (g2.east) -- (o2.west);
\node[prior, below=1mm of p2] (p3) {\textbf{ViVerBench}\\[-1pt] General visual verification};
\node[gap, right=4.0mm of p3] (g3) {Verification reliability across heterogeneous 3D evidence};
\node[ours, right=4.0mm of g3] (o3) {Ten 3D tasks evaluating decisions and reference-matched rejection explanations};
\draw[arrowleft] (p3.east) -- (g3.west);
\draw[arrowours] (g3.east) -- (o3.west);
\node[prior, below=1mm of p3] (p4) {\textbf{Eval3D}\\[-1pt] Asset consistency, alignment, and artifact localization};
\node[gap, right=4.0mm of p4] (g4) {Rejection-explanation correctness and verification of diverse, explicit 3D requirements};
\node[ours, right=4.0mm of g4] (o4) {Evaluates decisions across ten tasks including grasping and motion. Studies human-guided recursive refinement via explicit failure feedback};
\draw[arrowleft] (p4.east) -- (g4.west);
\draw[arrowours] (g4.east) -- (o4.west);
\end{tikzpicture}
\caption{\textbf{Evaluation scope relative to prior work.} Prior work~\citep{zhang2025_3dgenbench,zhang2025hi3deval,zhang2025omniverifier,duggal2025eval3d} emphasizes preference, hierarchical quality, general visual verification, or asset-level consistency and artifact localization. Our 3DVerifierBench targets complementary verifier-side questions through explicit requirements, controlled counterfactuals, heterogeneous 3D evidence, and verification-guided refinement.}
\label{fig:related_comparison}
\end{figure}
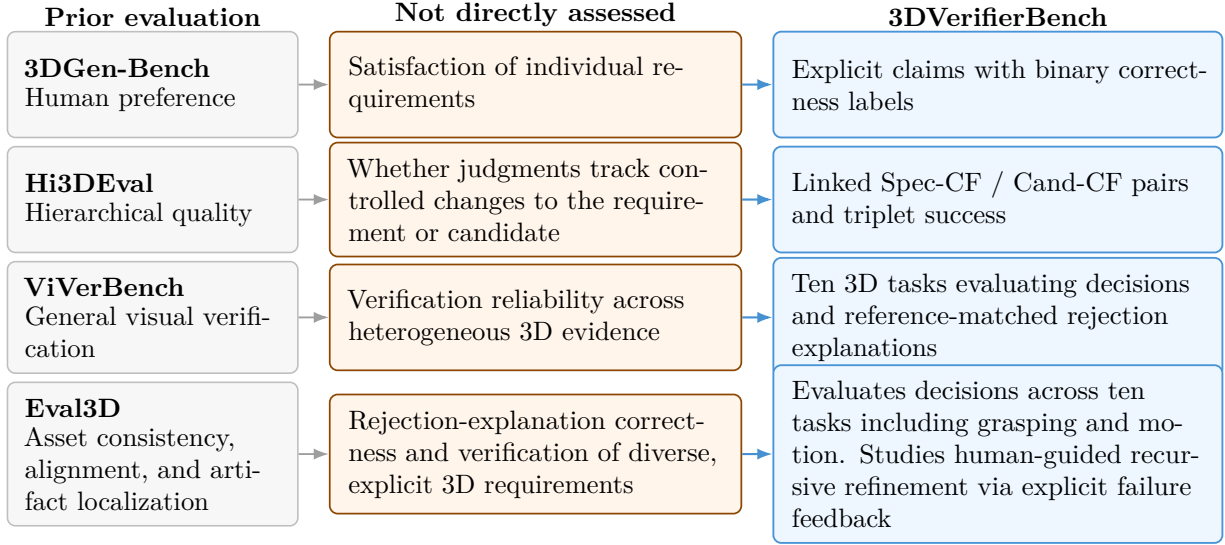

Recent 3D-language models connect native 3D representations to language and evaluate object, part, spatial, and embodied understanding \citep{xu2024pointllm,deng2025_3dllava,wang2025partnext,zhang2026embodied3dbench}. In parallel, 3D generation benchmarks and evaluators measure semantic alignment, human preference, geometric quality, and generation defects \citep{wu2024gpteval3d,zhang2025_3dgenbench,zhang2025hi3deval,jia2026db3dme,duggal2025eval3d,zhao2026_3ddefectbench}. Together, these works substantially improve our ability to describe and assess 3D content. However, predicting a property or scoring an asset does not directly measure whether a model can \textbf{verify a specific requirement}, reject a plausible but incorrect candidate, and identify heterogeneous 3D evidence that makes it wrong. We focus on this verifier-side evaluation against human judgments, with scope comparisons to prior work in Figure~\ref{fig:related_comparison}.

\section{3DVerifierBench}\label{dverifierbench}

To benchmark AI against human verification, 3DVerifierBench unifies 10 tasks under a common interface, evaluating reliability across decisions, explanations, and consistency. Together, these dimensions assess whether a verifier can provide correct, informative, and consistent feedback for reflection in 3D generation.

\begin{figure}[!htbp]
\captionsetup{font=small}
\centering

\includegraphics[width=\textwidth]{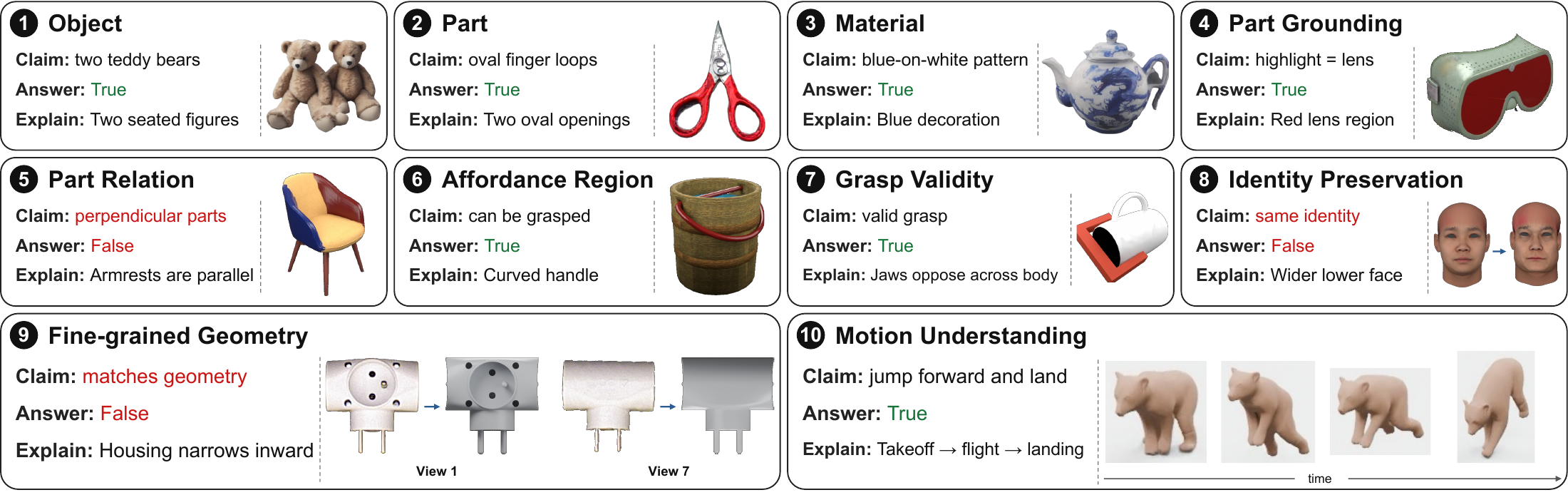}

{\small
\setlength{\tabcolsep}{4.5pt}
\resizebox{\linewidth}{!}{%
\begin{tabular}{@{}lcccccccccc@{}}
\toprule
Task & Object & Part & Mat. & Ground. & Relation & Aff. Reg. &
Grasp & Identity & FineGeo. & Motion \\
\midrule

Amount & 306 & 306 & 324 & 338 & 304 & 409 &
200 & 200 & 60 & 134 \\

\bottomrule
\end{tabular}}
}

\caption{\textbf{Overview of 3DVerifierBench.}
Each task presents task-specific 3D evidence together with a verification requirement, reference answer, and explanation. Evidence spans multi-view objects, marked parts and regions, grasp poses, identity comparisons, photo-to-geometry comparisons, and motion sequences. 
The table summarizes the number of examples and input evidence for each task, totaling 2,581 examples. Mat., Ground., Aff. Reg., and FineGeo. denote Material, Part Grounding, Affordance Region, and Fine-grained Geometry, respectively.
Views and text are abbreviated.}
\label{fig:tasks}
\end{figure}

\noindent\textbf{Unified Verification Interface and Task Taxonomy.}
Given task instructions \(I\), a claim \(C\), candidate evidence \(E^{\mathrm{cand}}\), and optional reference evidence \(E^{\mathrm{ref}}\), a verifier returns a binary answer \(\hat y\in\{\mathrm{True},\mathrm{False}\}\) and an explanation \(\hat e\):
$x=(I,C,E^{\mathrm{cand}},E^{\mathrm{ref}}), f_\theta(x)=(\hat y,\hat e).$
We denote \(\sigma=(C,E^{\mathrm{ref}})\) as the requirement, and the candidate may be an asset, part, grasp, or animation.
Verification checks whether 3D evidence fulfills requirements, rather than general plausibility or visual quality. For negative cases, reference explanations explicitly pinpoint the violated requirement and conflicting evidence.
Figure~\ref{fig:tasks} details the 3DVerifierBench architecture and statistics, which instantiates this common interface across complementary forms of 3D verification requirements.

\noindent\textbf{Answer, Explanation, and Consistency Metrics.}
We evaluate decisions, reasons, and responses to changes, and use Balanced Accuracy (BA) within each task and equal task weights overall.

For each task $t$, we report balanced accuracy,
$\textstyle \mathrm{BA}_t=\frac{1}{2}
(\mathrm{Acc}_{T,t}+\mathrm{Acc}_{F,t})$,
and aggregate performance with an overall accuracy 
$\textstyle \mathrm{OA}=\frac{1}{10}
\sum_{t=1}^{10}\mathrm{BA}_t$.
To distinguish a correct rejection from a supported explanation, a
reference-semantic Judge checks the explanations of correctly rejected negatives. Let $\mathrm{FalseJoint}_t$ denote joint answer-and-explanation success over all negatives. We define
$\textstyle \mathrm{ExpBA}_t=\frac{1}{2}
(\mathrm{Acc}_{T,t}+\mathrm{FalseJoint}_t)$
and
$\textstyle \mathrm{ExpOA}=\frac{1}{10}
\sum_{t=1}^{10}\mathrm{ExpBA}_t$.
The Judge receives the task, claim, and both explanations, without visual evidence or model identity; failure indicates a reference-semantic mismatch. We further perform human inspection to validate the Judge in Sec.~\ref{how-good-are-current-3d-verifiers}.

To test whether judgments track the requirement and candidate, we also evaluate linked counterfactuals. A pair contains a positive anchor and a related negative; a triplet contains the anchor and both specification-side and candidate-side negatives. All linked answers must be correct, and explanation-aware variants also check negative explanations. 
Appendix~\ref{app:C} defines the metrics and Appendix~\ref{e.5-seven-task-pair-and-triplet-consistency} reports the eligible comparison sets.

\FloatBarrier
\section{Dataset and Benchmark Construction}\label{dataset-construction}

To support reliable verification feedback, supervision should identify not only whether a candidate is wrong, but \textbf{which requirement is contradicted by the evidence}. We therefore combine existing 3D resources with controlled interventions and native supervision, using three complementary strategies to construct task-specific data. The pipeline is shown in Figure~\ref{fig:construction} with details provided in Appendix~\ref{app:B}.

\begin{figure}[!htbp]
\captionsetup{font=small}
\centering\includegraphics[width=0.96\textwidth]{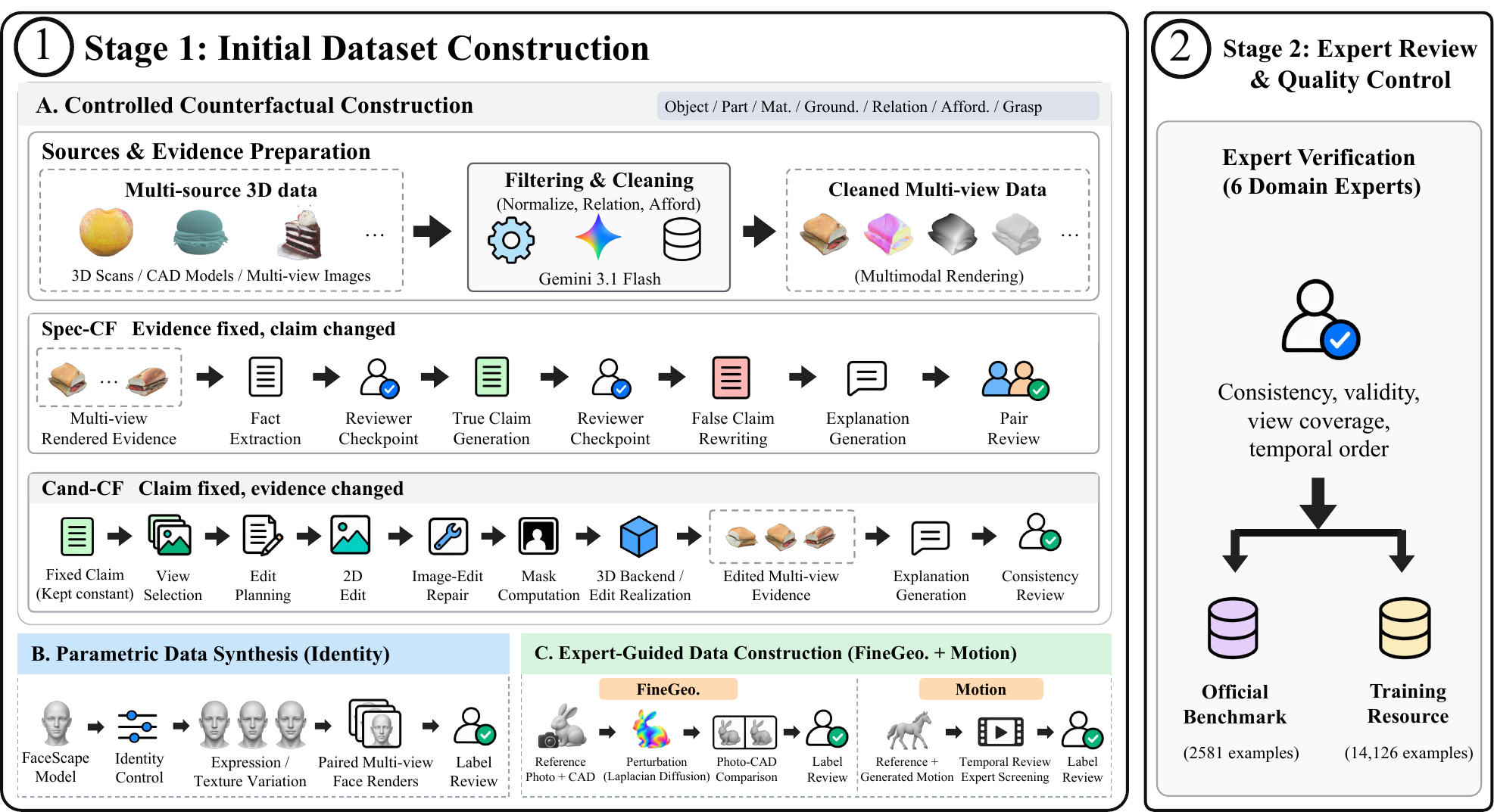}
\caption{\textbf{Our pipeline.} Reviewed evidence establishes positive anchors. Spec-CF changes the requirement under fixed evidence, whereas Cand-CF changes the candidate under the original requirement. Native annotations and task-specific evidence complement these controlled mutations. Automated checks, model-assisted review, and human review together produce the benchmark.}
\label{fig:construction}
\end{figure}

\noindent\textbf{Stage1: Initial Dataset Construction.}
We employ 3 strategies to construct data for the 10 tasks:

\begin{itemize}[leftmargin=*]
\item \noindent\textbf{Controlled Counterfactual Construction.}
For \{Object, Part, Mat., Ground., Relation, Aff.\ Reg., Grasp\}, we combine object-level assets and multi-view evidence from 3DGen-Bench, DB-3DME, and Hi3DBench~\citep{zhang2025_3dgenbench,jia2026db3dme,zhang2025hi3deval} with part annotations from PartNeXt and GAPartNet and grasp supervision
from ACRONYM~\citep{wang2025partnext,geng2023gapartnet,eppner2021acronym}. 
Starting from a positive anchor \((a,\sigma,\mathrm{True})\) where candidate \(a\) satisfies a reviewed specification \(\sigma\), we construct negative pairs via two counterfactual mechanisms:
Specification-side counterfactuals (\textbf{Spec-CF}) modify a key requirement while fixing \(a\), producing \((a,\sigma',\mathrm{False})\).
Candidate-side counterfactuals (\textbf{Cand-CF}) modify the asset, marked region, part pair, or contact while fixing the specification, producing \((a',\sigma,\mathrm{False})\).
Native annotations take precedence over synthetic edits when available, and candidate evaluation links are retained only when both records share identical model-visible claims. For Affordance Region, negative claims are verified by human review.
Two domain experts manually spot-check 10\% of the outputs at each
model-assisted review stage shown in Figure~\ref{fig:construction},
including the two reviewer checkpoints in Spec-CF, the final pair review
in Spec-CF, and the consistency review in Cand-CF.
\item \noindent\textbf{Parametric Data Synthesis.}
For \{Identity\}, we leverage FaceScape's bilinear model~\citep{yang2020facescape} to construct 200 balanced face pairs.
Specifically, identity variations are sampled using standardized identity coordinates, with labels assigned based on whether these identity coefficients are modified. To ensure that these parameter shifts yield human-perceptible identity differences, 6 domain experts were invited to establish explicit thresholds and validate the generated pairs.
To systematically evaluate robustness against non-identity attributes, \textit{i.e.}, \emph{expression and texture}, we vary them as nuisance factors across 4 conditions: (1) nearly gray texture with minimal changes,
(2) texture changes only,
(3) expression changes only, 
(4) changes to both.
\item \noindent\textbf{Expert-Guided Data Construction.}
For \{FineGeo., Motion\}, each has 6 domain experts guiding data construction and annotation.
For FineGeo., we apply Laplacian diffusion~\citep{desbrun1999implicit} to ground-truth T-LESS models~\citep{hodan2017tless}, with experts selecting surface regions and perturbation parameters.
Original and perturbed models yield positive and negative photograph-CAD comparisons against fixed reference photographs.
For Motion, we combine reference motions from AnimalML3D~\citep{yang2024omnimotiongpt} with sequences generated by AnimateAnyMesh~\citep{wu2025animateanymesh} and Animate3D~\citep{jiang2024animate3d}.
Experts classify motions as \emph{Correct}, \emph{Unnatural Motion},
\emph{Incorrect Motion}, or \emph{Deformation Artifacts}, retaining only samples with unanimous agreement.
\end{itemize}

\noindent\textbf{Stage2: Expert Review and Quality Control.}
We further invited 6 additional, non-overlapping domain experts to verify benchmark data.
Checks cover annotation consistency, perturbation bounds, mesh validity, view coverage, rendering completeness, and temporal order where applicable.
Source-specific preprocessing and validation details are in Appendix~\ref{app:B}.

\begin{table}[t]
\caption{
Results on 3DVerifierBench. 
All values represent the binary Balanced Accuracy, except for the 4-CLS column (four-class Exact Match accuracy).
Bold marks the best model score.
}
\label{tab:main}
\centering
\setlength{\tabcolsep}{3pt}
\resizebox{\textwidth}{!}{%
\begin{tabular}{@{}lccccccccccc|cc@{}}
\toprule

\multicolumn{1}{c}{\multirow[c]{2}{*}[-0.5ex]{\makecell[c]{Model /\\ Evaluator}}}
& \multirow[c]{2}{*}{Object}
& \multirow[c]{2}{*}{Part}
& \multirow[c]{2}{*}{Mat.}
& \multirow[c]{2}{*}{Ground.}
& \multirow[c]{2}{*}{Relation}
& \multirow[c]{2}{*}{Aff. Reg.}
& \multirow[c]{2}{*}{Grasp}
& \multirow[c]{2}{*}{Identity}
& \multirow[c]{2}{*}{FineGeo.}
& \multicolumn{2}{c|}{Motion}
& \multirow[c]{2}{*}{Overall}
& \multirow[c]{2}{*}{ExpOA} \\[-0.4ex]

\cmidrule(lr){11-12}

&
& & & & & & & &
& 4-CLS & Binary
& & \\

\midrule

\rowcolor{humanheader}
\multicolumn{14}{c}{\textbf{Human Performance}} \\
Human
& 95.0 & 82.5 & 90.0 & 92.5 & 95.0 & 90.0
& 80.0 & 92.9 & 80.0
& 70.0 & 80.0
& 87.8 & -- \\

\rowcolor{baselineheader}
\multicolumn{14}{c}{\textbf{Baselines}} \\
Random guess
& 50.0 & 50.0 & 50.0 & 50.0 & 50.0 & 50.0
& 50.0 & 50.0 & 50.0
& 25.0 & 50.0
& 50.0 & -- \\

\rowcolor{closedheader}
\multicolumn{14}{c}{\textbf{Proprietary MLLMs}} \\
GPT-6 Astra
& 75.3 & \textbf{72.1} & \textbf{71.5} & \textbf{89.6} & \textbf{90.3}
& 76.3 & 56.0 & 55.5 & 50.0
& \textbf{54.5} & \textbf{57.5}
& \textbf{69.4} & \textbf{65.8} \\

Gemini-3.8-Flash
& \textbf{76.5} & 69.1 & 70.4 & 85.5 & 87.7
& \textbf{77.3} & 54.5 & 57.0 & 50.0
& 40.3 & 56.0
& 68.4 & 57.8 \\

GPT-5.6-Sol
& 74.5 & 69.1 & 66.2 & 84.4 & 80.2
& 62.7 & 53.5 & 50.5 & 51.7
& 50.7 & 56.7
& 64.9 & 59.6 \\

\rowcolor{openheader}
\multicolumn{14}{c}{\textbf{Open-Source MLLMs}} \\
Gemma-4-31B
& 76.2 & 62.3 & 69.9 & 74.5 & 72.3
& 52.8 & 59.0 & 58.5 & 45.0
& 37.3 & 56.0
& 62.6 & 52.2 \\

Qwen3.5-122B-A10B
& 70.6 & 64.7 & 70.8 & 75.7 & 67.6
& 52.0 & 58.0 & 52.5 & \textbf{53.3}
& 26.9 & 50.7
& 61.6 & 48.9 \\

Qwen3.6-35B-A3B
& 62.3 & 59.8 & 60.0 & 68.7 & 64.8
& 49.5 & \textbf{66.5} & \textbf{59.0} & \textbf{53.3}
& 36.6 & 50.7
& 59.5 & 51.3 \\

InternVL3.5-38B
& 66.2 & 57.8 & 63.4 & 54.5 & 57.8
& 56.2 & 59.0 & 49.5 & 50.0
& 26.1 & 52.2
& 56.7 & 44.5 \\

GLM-4.6V-106B
& 66.7 & 58.1 & 62.0 & 57.0 & 54.2
& 50.0 & 50.0 & 49.5 & 50.0
& 44.0 & 53.0
& 55.0 & 50.9 \\

OmniVerifier-7B
& 58.3 & 53.7 & 53.7 & 50.0 & 46.6
& 52.7 & 51.5 & 50.0 & 51.7
& 21.6 & 52.2
& 52.0 & 30.0 \\

Qwen2.5-VL-7B-Instruct
& 54.4 & 51.2 & 50.9 & 50.0 & 50.0
& 50.4 & 50.0 & 50.0 & 45.0
& 16.4 & 50.0
& 50.2 & 22.1 \\

\bottomrule
\end{tabular}}
\end{table}

\subsection{Native Supervision and Quality Control}\label{native-supervision-and-quality-control}

When native labels already define the relevant distinction, we use them directly instead of forcing every task through the same editing backend. Part and relation annotations, action semantics, grasp outcomes, identity parameters, reference photographs, and reviewed motion sequences provide task-specific evidence for the corresponding claims. This keeps the verification interface unified while allowing the underlying evidence to match the semantics of each task.

Benchmark examples undergo task-specific human review for label correctness,
evidence sufficiency, and explanation quality. Excluding benchmark source groups
and their descendants from the 17,223-example construction pool yields a
\textbf{14,126-example Training Resource} with 4,011 source groups and zero
benchmark--training source-group overlap (Appendix~\ref{app:A}).

\section{Experiments}\label{experiments}

\subsection{Setup}\label{setup}

We evaluate GPT-6 Astra \citep{openai2026astra},
GPT-5.6-Sol \citep{openai2026gpt56},
Gemini-3.8-Flash \citep{google2026gemini38},
Qwen3.5-122B-A10B and Qwen3.6-35B-A3B \citep{qwen2026qwen35,qwen2026qwen36},
GLM-4.6V-106B \citep{zai2025glm46v},
Gemma-4-31B \citep{google2026gemma4},
InternVL3.5-38B \citep{wang2025internvl35},
Qwen2.5-VL-7B-Instruct \citep{bai2025qwen25vl},
and OmniVerifier-7B \citep{zhang2025omniverifier}
on all 2,581 examples. Direct inference uses temperature 0, top-\(p\) 1, and a 512-token output limit. Static evidence is fixed across models. Human scores use a 400-example subset with 40 examples per task. Confidence intervals resample task-specific clusters, retaining linked examples together and using paired draws for human-model differences (Appendix~\ref{app:C}).
Motion Understanding uses native MP4 when supported and a sixteen-frame sheet otherwise (Appendix~\ref{app:D}). 

\subsection{How Good Are Current MLLMs in 3D Verification?}\label{how-good-are-current-3d-verifiers}

Table~\ref{tab:main} shows the experimental results on 3DVerifierBench. GPT-6 Astra achieves the highest Overall score of 69.4\%, followed by Gemini-3.8-Flash at 68.4\%. Astra reaches 90.3\% on Part Relation and 89.6\% on Part Grounding,
but only 55.5\% on Identity and 50.0\% on Fine-grained Geometry (FineGeo.). These weaknesses extend across models: None exceeds 59.0\% on Identity or 53.3\% on Fine-grained Geometry.
We also provide an analysis of human performance on its subset, which is detailed in Appendix \ref{app:G}. We observe that the performance of existing MLLMs is still far behind humans. For example, humans achieve an 87.8\% Overall score versus Astra's 66.2\%. Note that this result differs from the one in Table 1 due to different testing data.

Useful feedback requires correct rejections and relevant reasons. Under reference-semantic explanation checking, ExpOA falls below Overall score by 3.6 to 28.1 points across models. Astra remains strongest at 65.8\% (95\% CI: [64.4, 67.2]\%). GPT-5.6-Sol overtakes Gemini-3.8-Flash despite its lower answer score. Across all models, 36.5\% of 9,488 correct rejections fail the reference-semantic Judge (Appendix~\ref{app:H}).

\begin{table}[htbp]
\caption{
Human validation of the explanation Judge.
We sample 30 correctly rejected model responses per task.
The numerical value in the bottom row is the Agreement rate (\%), which represents the ratio of identical Judge and human assessments of explanation validity.
}
\label{tab:judge_human}
\centering

\resizebox{\textwidth}{!}{%
\begin{tabular}{@{}ccccccccccc@{}}
\toprule
Material & Object & Part & Aff. Reg. & Grasp & Ground. & Relation & Identity & FineGeo. & Motion & Overall \\
\midrule

93.3 & 90.0 & 93.3 & 83.3 & 86.7 & 86.7 & 100.0 & 93.3 & 100.0 & 83.3 & 91.0 \\
\bottomrule
\end{tabular}
}

\end{table}

We additionally validate the explanation Judge against independent human assessment, as shown in Table~\ref{tab:judge_human}. We sample 30 correctly rejected responses per task, totaling 300 explanations. The Judge agrees with human assessment by \textbf{91.0\%}; 6.0\% are accepted by the Judge but rejected by humans, while 3.0\% are rejected by the Judge but accepted by humans. These results support the use of the Judge for aggregate explanation analysis, while the targeted
audits in Appendix~\ref{app:H} examine specific geometric and deformation failures.

We also provide controlled experiments to test whether judgments respond correctly to changes in requirements or candidates, as detailed in Appendix~\ref{e.5-seven-task-pair-and-triplet-consistency}. On 615 human-validated triplets, Gemini-3.8-Flash has the highest answer-only success at 49.6\%, while Astra reaches 46.5\% and leads the explanation-aware score at 40.5\% (Table~\ref{tab:e10}). These results test consistency within linked comparisons, complementing the Overall score.

\subsection{Limitations of MLLMs in 3D Verification}\label{what-limits-reliable-3d-verification}

\begin{figure}[!htbp]
\centering\includegraphics[width=1\textwidth]{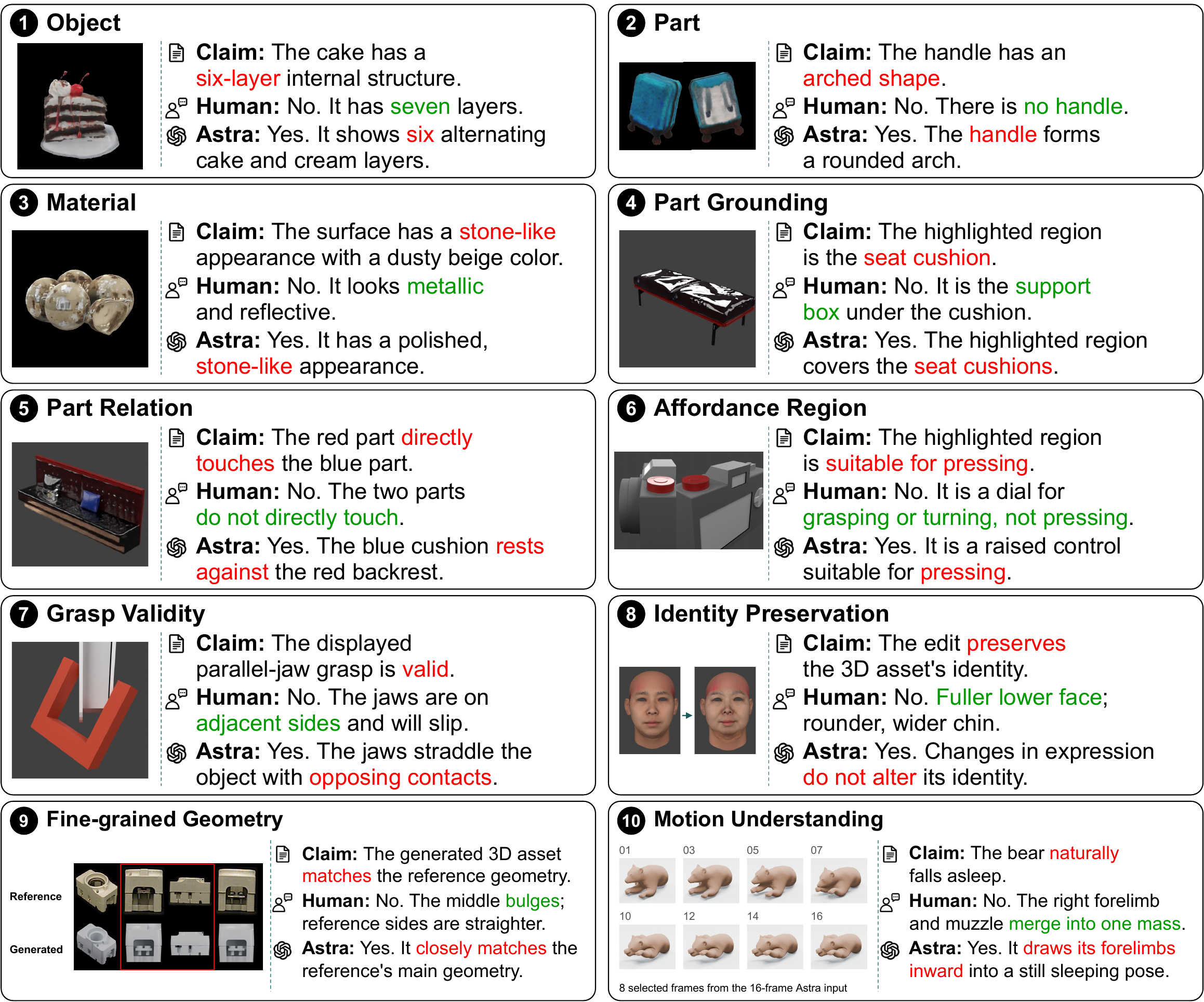}

\begin{findingbox}
\small
\noindent\textbf{Findings: Limitations of MLLMs in 3D Verification}
\begin{enumerate}[
  leftmargin=1.5em,
  topsep=4pt,
  itemsep=2pt,
  parsep=0pt
]
  \itshape
  \item Repeated structures miscounted.
  \item Nonexistent parts hallucinated from claim premises.
  \item Material descriptions accepted despite conflicting reflectance cues.
  \item Highlighted regions assigned incorrect semantic part labels.
  \item Direct contact inferred without checking actual spatial relations.
  \item Action-specific affordances confused across similar controls.
  \item Adjacent grasp contacts mistaken for opposing contacts.
  \item \textbf{Identity-altering shape changes mistaken for expression changes.}
  \item \textbf{Local curvature and contour mismatches are overlooked despite multi-view evidence.}
  \item \textbf{Deformation artifacts in motion sequences go undetected}.
\end{enumerate}
\end{findingbox}
\caption{Typical GPT-6 Astra failure cases and findings of limitations of MLLMs in 3D verification. 
}
\label{fig:failures}
\end{figure}

To identify concrete bottlenecks in reliable verification feedback, we present representative failures in Figure~\ref{fig:failures}. These examples reveal errors in fine-grained geometry, spatiotemporal validity, and the grounding of claims in visual evidence. We summarize them as ten failure modes in the findings below. Appendix~\ref{app:H} further provides details of category-wise quantitative analysis of these verification failures.

\subsection{How Does Evidence Representation Affect Verification?}\label{motion-evidence-ablation}

\noindent\textbf{Multi-view provides a practical protocol for 3D verification.} Multi-view representation generally performs better than raw 3D input, indicating that rendering 3D evidence into views is currently a more effective way to elicit reliable judgments from existing MLLMs. We compare multi-view RGB and raw OBJ input on matched Object, Part, and Material subsets in Table~\ref{tab:evidence_representation}(a). The multi-view results are also broadly consistent with the corresponding results in Table~\ref{tab:main}. This choice preserves the core difficulty of 3D verification, requiring models to aggregate partial observations, resolve occlusions, associate parts across views, and disentangle underlying geometry from viewpoint or appearance changes. As an early benchmark for 3D verifiers, 3DVerifierBench therefore adopts multi-view rendering as a practical and extensible protocol for grounding verification in 3D content.

Native point-cloud LLMs show strong response bias and unstable output
formatting under our verifier interface; we therefore report them only
in Appendix~\ref{e.4-native-3d-baselines}.

\noindent\textbf{For motion, no single representation is always the best.} Motion verification requires both temporal dynamics and geometric evidence. We evaluate three representations on the Motion Understanding task in Table~\ref{tab:evidence_representation}(b): 16 uniformly sampled frames, raw 3D motion in .obj sequences, and native video. We report binary and 4-class Accuracy.
No single representation consistently outperforms others across all models and metrics. \textbf{Native video} is the highest in binary (59.7\%), whereas \textbf{16-frames} yield the highest in 4-class (43.8\%). The preferred representation also varies by model: Gemini-3.8-Flash benefits most from native video, whereas Qwen3.6-35B-A3B excels with .obj sequences. These results show that denser temporal evidence alone does not guarantee superior motion verification; rather, performance depends on a model's ability to extract and integrate spatio-temporal cues. Consequently, we adopt native video / 16-frame sampling as the default settings of our benchmark.

\begin{table}[htbp]
\caption{
Evidence representation ablations.
(a) Multi-view RGB \textit{v.s.} raw 3D input.
Each cell reports multi-view / raw OBJ Balanced Accuracy (\%).
(b) Motion verification under different evidence representations.
Each cell reports binary / 4-class Acc.
}
\label{tab:evidence_representation}

\centering
\scriptsize

\begin{minipage}[t]{0.49\linewidth}
\centering
\textbf{(a) Multi-view v.s. raw 3D}\\[2pt]

\setlength{\tabcolsep}{2pt}
\resizebox{\linewidth}{!}{%
\begin{tabular}{@{}lrrrr@{}}
\toprule
Model & Object & Part & Material & Avg. \\
\midrule

Gemini-3.8
& 62.50 / 56.25
& 68.75 / 18.75
& 75.00 / 56.25
& 68.75 / 43.75 \\

Gemma-4
& 62.50 / 37.50
& 62.50 / 50.00
& 68.75 / 37.50
& 64.58 / 41.67 \\

Qwen3.5
& 68.75 / 50.00
& 62.50 / 50.00
& 62.50 / 50.00
& 64.58 / 50.00 \\

Qwen3.6
& 56.25 / 50.00
& 68.75 / 50.00
& 68.75 / 50.00
& 64.58 / 50.00 \\

\midrule
Mean
& 62.50 / 48.44
& 65.63 / 42.19
& 68.75 / 48.44
& 65.63 / 46.35 \\

\bottomrule
\end{tabular}}
\end{minipage}
\hfill
\begin{minipage}[t]{0.49\linewidth}
\centering
\textbf{(b) Motion: frames v.s. 3D seq. v.s. video}\\[2pt]

\setlength{\tabcolsep}{3pt}
\begin{tabular*}{\linewidth}{
@{\extracolsep{\fill}}
lccc
@{}
}
\toprule
Model & Frames & 3D OBJ seq. & Native Video \\
\midrule

Gemini-3.8
& 65.3 / \textbf{56.2}
& 49.4 / 40.6
& \textbf{68.2} / 50.0 \\

Gemma-4
& 49.4 / 43.8
& \textbf{63.5} / \textbf{46.9}
& \textbf{63.5} / 43.8 \\

Qwen3.5
& \textbf{55.1} / 31.2
& 48.8 / 25.0
& 54.7 / \textbf{34.4} \\

Qwen3.6
& 56.9 / 43.8
& \textbf{58.2} / \textbf{50.0}
& 52.4 / 40.6 \\

\midrule
Mean
& 56.7 / \textbf{43.8}
& 55.0 / 40.6
& \textbf{59.7} / 42.2 \\

\bottomrule
\end{tabular*}
\end{minipage}

\end{table}

\subsection{How Can Verification Feedback Help 3D Generation Models Reflect?}\label{can-verification-improve-3d-generation}

We examine how reliable verification feedback can help 3D systems reflect on their outputs. Parallel selection measures the value of recognizing successful candidates, while human-guided recursive refinement examines how explicit failure feedback can guide subsequent revisions.

\textbf{Parallel selection.} 
For 40 source images, we fix seven factual requirements before generation and obtain eight candidates each from Pixal3D and Hunyuan3D-2.1~\citep{hunyuan3d21}, forming a sixteen-candidate pool per input. All candidates are reviewed by multiple humans with adjudication.
Table~\ref{tab:verification_generation} shows that human-oracle selection raises full factual success from 26.7\% under uniform random selection to 90.0\%, while Strict Full rises from 16.6\% to 85.0\%.
Selection by GPT-6 Astra
also improves over random, approaches human-level performance in factual success rate, but stays below the human oracle.
Thus, the oracle results quantify the \emph{selection headroom} available from reliable verification rather than the performance of an existing automatic verifier.

\begin{table}[htbp]
\caption{What reliable verification could enable in 3D generation.
Left: candidate selection from sixteen-candidate pools over 40 inputs.
Right: human-guided recursive refinement on eight inputs.
Full requires all seven facts; Strict additionally requires no severe defects and source fidelity.}
\label{tab:verification_generation}
\centering
\small
\setlength{\tabcolsep}{4pt}
\renewcommand{\arraystretch}{1.05}

\begin{minipage}[t]{0.48\linewidth}
\centering
\textbf{Parallel Selection} \\[2pt]
\begin{tabular}{@{}lccc@{}}
\toprule
Selector & Fact Sat. & Full & Strict \\
\midrule
Random       & 80.8 & 26.7 & 16.6 \\
GPT-6 Astra  & 90.4 & 55.0 & 45.0 \\
Human Oracle & \textbf{98.2} & \textbf{90.0} & \textbf{85.0} \\
\bottomrule
\end{tabular}
\end{minipage}
\hfill
\begin{minipage}[t]{0.48\linewidth}
\centering
\textbf{Recursive Refinement} \\[2pt]
\begin{tabular}{@{}lc@{}}
\toprule
Setting & Full Success \\
\midrule
Parallel Oracle@8     & 3/8 (37.5) \\
Recursive             & \textbf{8/8 (100.0)} \\
\quad w. no severe defect & 7/8 (87.5) \\
\quad w. source fidelity  & 6/8 (75.0) \\
\bottomrule
\end{tabular}
\end{minipage}
\end{table}

\textbf{Recursive refinement.} 
Selection cannot recover a requirement that is absent from all sampled candidates.
We therefore study eight inputs whose initial Hunyuan3D-2.1 generation violates at least one of the seven facts.
After each generation, a human verifier identifies unmet requirements and guides either regeneration from a revised 2D condition or targeted mesh editing
with VecSet-Edit~\citep{hsiao2026vecsetedit}.
Recursive refinement satisfies all seven facts for all $8/8$ cases, although only $7/8$ final assets avoid severe defects and $6/8$ additionally preserve source fidelity.
We additionally compare against a parallel oracle baseline, Parallel Oracle@8, which uses a fixed pool of eight independently generated candidates and selects a successful candidate whenever one exists. 
In contrast, it succeeds on only $3/8$ cases (37.5\%).
These human-guided results suggest that repeated feedback and revision can satisfy requirements not met by parallel candidates, while also showing that factual success alone does not guarantee asset quality or source fidelity.

\begin{figure}[!htbp]
\captionsetup{font=small}
\centering\includegraphics[width=0.92\textwidth]{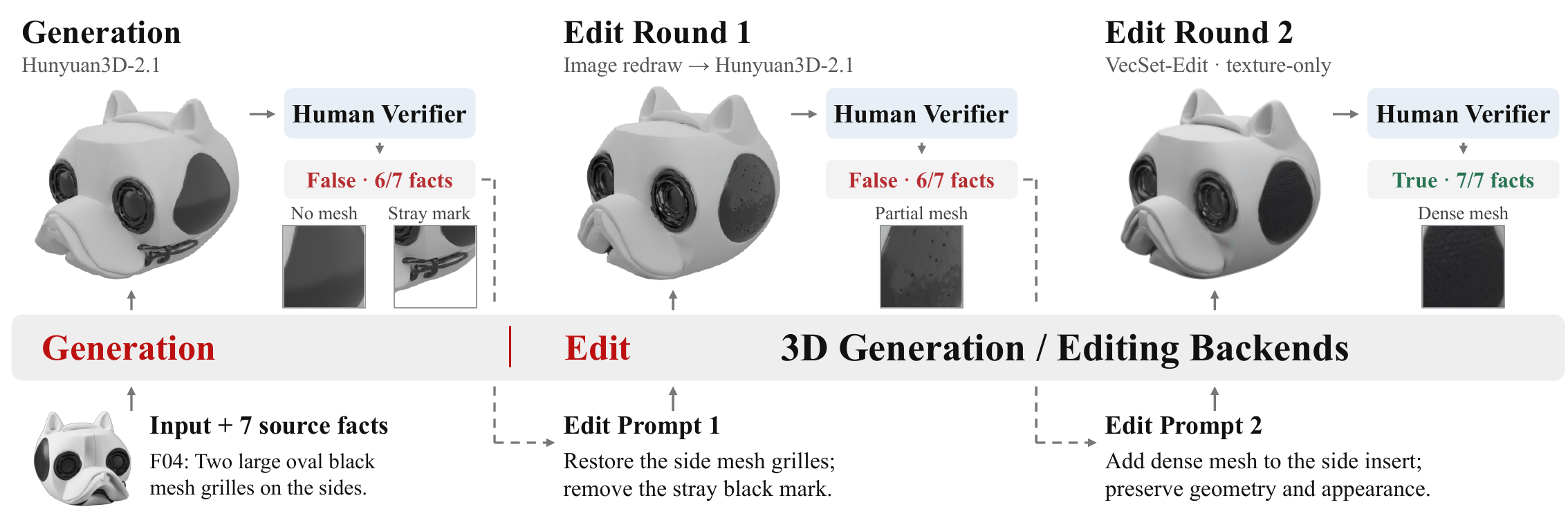}
\caption{Human-guided 3D recursive refinement trajectory. Given the source image and target facts, initial generation exhibits a missing side grille and a surface artifact (left, 6/7 facts). A revised one eliminates the artifact but leaves the grille incomplete (middle, 6/7 facts). Subsequent edit fully restores the mesh appearance and satisfies all facts (right, 7/7 facts). Intermediate 2D conditions and masks are omitted.}
\label{fig:refinement}
\end{figure}

Figure~\ref{fig:refinement} illustrates one trajectory in which repeated verification identifies an unresolved grille requirement and guides a subsequent targeted correction.

\par\noindent\begin{minipage}{\linewidth}
\setlength{\parskip}{4pt}
\section{Discussion and Conclusion}\label{discussion-and-conclusion}

To reflect on a 3D output, a system must determine whether it meets its requirements and explain any violations. 3DVerifierBench shows that current models often struggle to provide this feedback, even when generated content looks plausible. Answer, explanation, and counterfactual-consistency evaluations reveal failures to identify decisive facts, their targets, and relevant contradictions. Our construction pipeline provides structured data to support training on these distinctions. Human-oracle selection and human-guided refinement illustrate how better feedback can help select and revise 3D assets, although satisfying explicit facts does not ensure overall quality or source fidelity. These findings motivate more reliable evidence-grounded feedback for future 3D generation systems.
\end{minipage}\par

\clearpage
\subsection*{AI use statement}

In this work, we used generative AI tools for research ideation and
execution, including assistance with research methodology, experimental
design, implementation, and the generation of synthetic data.
In particular, generative AI tools were used in parts of our
model-assisted data construction pipeline, including multi-view fact
extraction, verification-claim construction and review, and the
generation and editing of counterfactual candidates.
We have not used generative AI tools for formulating or proving
mathematical claims, and other required-disclosure tasks that are not
part of this work are not applicable.
Additionally, we used generative AI tools to aid or polish writing and
for retrieval and discovery, including identifying potentially relevant
prior work.
We have reviewed all AI-assisted work. AI-assisted data construction
was subject to automated checks, model-assisted review, and human
review, and experimental outputs and analyses used in the paper were
checked by the authors. We take responsibility for the final content of
this work, including text, claims, data, code, analyses, or artifacts
produced with the aid of generative AI.

\subsection*{Ethics statement}
For facial geometry, we use the FaceScape bilinear statistical model rather than participant-level scans. The model compresses the identity variation of 847 subjects into a 50-dimensional PCA space, from which we synthesize facial geometry. We do not use or redistribute the original FaceScape images, textures, subject-level meshes, demographic metadata, or identity-specific records.

\bibliography{references,reflection_references}
\bibliographystyle{plainnat}

\clearpage
\appendix
\numberwithin{equation}{section}
\numberwithin{table}{section}
\renewcommand{\thetable}{\thesection\arabic{table}}
\numberwithin{figure}{section}
\renewcommand{\thefigure}{\thesection\arabic{figure}}
 \input{appendix.tex}

\end{document}

%% file: math_commands.tex
\usepackage{amsmath,amsfonts,bm}

\def\eqref#1{equation~\ref{#1}}

\def\1{\bm{1}}

\DeclareMathAlphabet{\mathsfit}{\encodingdefault}{\sfdefault}{m}{sl}
\SetMathAlphabet{\mathsfit}{bold}{\encodingdefault}{\sfdefault}{bx}{n}

%% file: appendix.tex
\section{Tasks and Data Composition}\label{app:A}\label{appendix-a.-tasks-and-data-composition}

\subsection{Task Definitions}\label{a.1-task-definitions}

Object checks identity, global shape, and object-level properties.
Part checks component presence, count, type, shape, and arrangement.
Material checks color, texture, patterns, and visible surface appearance.
Part Grounding presents a highlighted region and a named-part claim.
Part Relation presents marked subject and object parts and a relation claim;
the markings define its direction. Affordance Region checks whether a marked
region supports a stated action. Grasp Validity displays a parallel-jaw pose
relative to an object and uses source simulation labels.

Identity Preservation asks whether two multi-view assets retain the same identity despite changes in expression or texture. Fine-grained Geometry asks whether a generated 3D candidate matches the reference object's shape and structure, while explicitly ignoring differences in color, texture, and material. Motion Understanding evaluates the complete caption set and checks action correctness, naturalness, and geometric stability. Its three negative issue classes are internal diagnostics, not additional tasks.

A \emph{counterfactual triplet} contains a positive anchor, a
specification-side negative, and a candidate-side negative. An anchor with only
one available negative forms a pair rather than a complete triplet.

A counterexample is not automatically a measured counterfactual pair. The latter requires the original and changed records, their shared specification or candidate, and the sibling link. An A/B identity comparison is likewise one verification input, not two independently scored pair/triplet branches.

\subsection{Model-Visible Inputs}\label{a.2-model-visible-inputs}

Object, Part, and Material receive the claim and eight frozen original views, without an additional task-specific instruction. Part Grounding, Part Relation, Affordance Region, and Grasp Validity receive their task-specific instructions and evidence sheets. Identity uses one 2-by-8 asset-pair sheet. Fine-grained Geometry uses one 2-by-7 reference/render sheet. Motion Understanding uses native MP4 or a frozen 4-by-4 sheet of sixteen uniformly sampled frames. A video and all its captions form one example. Reference answers, reference explanations, construction labels, and analysis labels are excluded from subject-model inputs.

\subsection{Benchmark and Training Resource}\label{a.3-benchmark-and-construction-pool}\label{a.3-benchmark-and-training-resource}

The benchmark contains 2,581 examples: 970 True and 1,611 False. All ten models evaluate this set; the human comparison uses forty examples per task drawn from it.

Starting from the 17,223-example automated construction pool, source-level grouping and exclusion of all benchmark source groups and their descendants yield a source-disjoint \textbf{Training Resource} with 14,126 examples. It contains 5,660 True anchors and 8,466 False examples in 4,011 source groups. Table~\ref{tab:a3} reports the task-wise composition.

\begin{table}[htbp]
\caption{Source-disjoint Training Resource obtained from the 17,223-example construction pool after excluding benchmark source groups and their descendants. Per-task source-group counts are not additive because a source group can contribute to multiple tasks.}
\label{tab:a3}
\centering\setlength{\tabcolsep}{3pt}
\begin{adjustbox}{max width=\linewidth}
\begin{tabular}{@{}lrrrrr@{}}
\toprule
Task & Examples & True & False & Anchors & Source groups \\
\midrule
Object & 3,912 & 1,654 & 2,258 & 1,654 & 1,650 \\
Part & 3,284 & 1,331 & 1,953 & 1,331 & 1,329 \\
Material & 1,846 & 811 & 1,035 & 811 & 811 \\
Part Grounding & 1,424 & 477 & 947 & 477 & 477 \\
Part Relation & 1,392 & 464 & 928 & 464 & 464 \\
Affordance Region & 1,291 & 434 & 857 & 434 & 434 \\
Grasp Validity & 977 & 489 & 488 & 489 & 484 \\
\midrule
Total & 14,126 & 5,660 & 8,466 & 5,660 & 4,011 \\
\bottomrule
\end{tabular}
\end{adjustbox}
\end{table}

\subsection{Source Grouping}\label{a.4-source-grouping}

All examples derived from the same source asset are assigned to the same source group, including counterfactual branches and edited descendants; known duplicates across source collections are merged before partitioning. Grouping uses source identities together with branch/descendant relations, declared asset hashes, path aliases, source-view hashes, and known upstream identities. The release contains 1,123 benchmark source groups and 4,011 training source groups, with zero source-group overlap. Source asset keys, declared asset hashes, and known-duplicate groups also have zero benchmark--training overlap. Per-task source-group counts in Table~\ref{tab:a3} are not additive because a source group can contribute to multiple tasks.

\section{Construction and Review}\label{app:B}\label{appendix-b.-construction-and-review}

\subsection{Source Preparation}\label{b.1-source-preparation}

Mesh preparation checks file validity, coordinate normalization, ground placement, and renderability. Fixed cameras and lighting produce consistent evidence. Screening checks completeness and factual visibility; minor defects remain acceptable when the relevant fact is clear. DB-3DME supplies eight horizontal turntable observations. It has no editable source mesh for Material changes. PartNeXt checks cover face-to-part mappings; GAPartNet checks cover actionable regions and articulations; ACRONYM preparation aligns meshes and gripper poses.

\textbf{Identity.} We use FaceScape's bilinear face model \citep{yang2020facescape}, retaining identity and expression coefficients as construction metadata. Each A/B comparison uses eight corresponding cameras, with the reference in the upper row and candidate in the lower row. The 2-by-8 RGB sheet is 1536-by-384 pixels; its constituent renders are produced at 512-by-512. Model inputs exclude parameter values, construction conditions, and labels.
The full text prompt for evaluating MLLMs' verification in this task is:
\begin{textpromptbox}
{\footnotesize\sffamily\color{black!50} Identity Text Prompt}\par
\smallskip
We edited the 3D asset. Please evaluate whether the edit preserves its identity.
\end{textpromptbox}

\textbf{Fine-grained Geometry.} T-LESS provides thirty source objects, reference photographs, and corresponding CAD geometry \citep{hodan2017tless}. These CAD models are source assets, not candidate code: No program is presented to the verifier. We align seven reference and render directions and stack them in a 3728-by-1120 sheet. Each object's positive and negative branches reuse the same upper-row photographs. 
The full text prompt for evaluating MLLMs' verification in this task is:
\begin{textpromptbox}
{\footnotesize\sffamily\color{black!50} Fine-grained Geometry Text Prompt}\par
\smallskip
You are evaluating a generated 3D asset. Given reference images and rendered views of the generated asset, decide whether the candidate matches the reference object's shape and structure. Differences in color, texture, material should not affect your judgment. Answer Y/N.
\end{textpromptbox}

\textbf{Motion Understanding.} The source collection contains 134 animation examples constructed using reference motions from AnimalML3D~\citep{yang2024omnimotiongpt}, with candidate sequences generated by AnimateAnyMesh and Animate3D. Each generated MP4 and its associated set of three to five captions are retained as one example. The videos use 24 fps and contain 13 to 151 frames. Native-video inputs retain the MP4. Image inputs use sixteen frames at indices \(\mathrm{round}(k(F-1)/15)\) for \(k=0,\ldots,15\), arranged chronologically in a 4-by-4 sheet; short videos may repeat frames. The input contains neither source failure-folder names nor reference labels.
The full text prompt for evaluating MLLMs' verification in this task is:
\begin{textpromptbox}
{\footnotesize\sffamily\color{black!50} Motion Understanding Text Prompt}\par
\smallskip
Determine whether the 3D motion matches the intended motion described by the following captions.
The captions describe different aspects or stages of the same intended motion. Consider them together:

1. \{caption\_1\}

2. \{caption\_2\}

3. \{caption\_3\}

Answer Yes or No. If No, specify one issue: Unnatural motion, Incorrect motion, or Deformation artifacts.
\end{textpromptbox}

\textbf{Explanation Judge.}
The explanation Judge compares a model-generated explanation with the trusted
reference explanation for the same task and claim.
The full text prompt used by the explanation Judge is:
\begin{textpromptbox}
{\footnotesize\sffamily\color{black!50} Explanation Judge Text Prompt}\par
\smallskip
You evaluate whether a verifier explanation is semantically correct against a
trusted reference explanation. Do not re-solve the visual task and do not
require wording overlap.

The candidate explanation is valid exactly when it identifies the same
violated fact and contradiction as the reference explanation. Accept concise
paraphrases and equivalent descriptions. It is invalid if it changes the
count, value, relation, material, region, or direction; discusses the wrong
fact; adds a conflicting hallucination; or merely says the claim is false
without identifying the contradiction.

The reference explanation is a trusted evaluation annotation.

\smallskip
Task: \{task\_id\}

Claim: \{claim\}

Candidate explanation: \{model\_explanation\}

Reference explanation: \{reference\_explanation\}
\end{textpromptbox}

The full system prompt used by the verification evaluation %for Object, Part, Material 
is:
\begin{textpromptbox}
{\footnotesize\sffamily\color{black!50} Verification System Prompt}\par
\smallskip
You verify factual claims about a 3D object from the supplied visual evidence. Judge only what is supported by the supplied public input. Return exactly one JSON object and no markdown: \{``answer'': true\_or\_false, ``explanation'': "brief factual reason"\}. The answer value must be a JSON boolean, not a string.
\end{textpromptbox}

\subsection{Facts, Anchors, and Spec-CF}\label{b.2-facts-anchors-and-spec-cf}

Fact extraction covers Object, Part, and Material. RGB provides semantic evidence. Additional depth, normal, or shading information can support construction-stage boundary checks. A writer combines compatible facts into a True Anchor. Review checks the claim against the source evidence.

Spec-CF changes a count, color, shape, direction, or part property. Other entities and the claim's scope are retained. Pair review checks the original and rewritten claims together. The negative reference identifies the changed requirement and the candidate's actual property.

\subsection{Cand-CF: View Selection, Masks, and 3D Realization}\label{b.3-cand-cf-view-selection-masks-and-3d-realization}

View selection considers target visibility, occlusion, surrounding structure, and backend requirements. The selected conditioning view exposes the intended change. The planner records the target, non-target preservation constraints, mask mode, and generation route. The benchmark retains its fixed evaluation views.

Image editing uses a local mask, a full-object mask, or an unmasked instruction. FLUX.2 [klein] 9B~\citep{bfl2026flux2klein} produces an initial edit. Review checks the target change, object integrity, boundaries, and preservation. Qwen-Image-Edit-2511~\citep{wu2025qwenimage} applies revisions from this feedback.

After the image edit is accepted, source and edited images are registered. Image differences, connected regions, and boundary checks refine a posterior change mask. This mask records the realized change and is separate from the conditioning mask specifying the intended edit.

\begin{table}[htbp]
\caption{Image-mediated 3D realization for object attributes.}
\label{tab:b1}
\centering
\begin{adjustbox}{max width=\linewidth}
\begin{tabular}{@{}lrr@{}}
\toprule
Level & Route & Acceptance checks \\
\midrule
Object & Pixal3D reconstruction & Global change, identity, required structure \\
Part & Pixal3D and VecSet-Edit & Target part, connectivity, non-target preservation \\
Material & VecSet-Edit & Visible appearance, required geometry \\
\bottomrule
\end{tabular}
\end{adjustbox}
\end{table}

Part construction compares candidates from both routes. Final assets are normalized and rendered again. Review checks that the original anchor remains true on the source and becomes false on the changed candidate, with an observable contradiction. The negative explanation describes the changed asset rather than the requested edit alone.

\subsection{Human Review}\label{b.5-human-review}

During model-assisted construction, two domain experts manually spot-check
10\% of the outputs at each review checkpoint shown in
Figure~\ref{fig:construction}. These checkpoints include the two Spec-CF
reviewer checkpoints, Spec-CF pair review, and Cand-CF consistency review. 

After construction, six additional domain experts, non-overlapping with
the construction-stage reviewers, independently verify the benchmark
candidates. They check label correctness, evidence sufficiency,
perturbation validity, mesh and rendering quality, view coverage, and,
for Motion Understanding, temporal consistency. Only samples that pass
this expert review are retained in the benchmark. Pair and triplet
inclusion additionally requires the fixed-evidence or fixed-claim
conditions in Appendix~\ref{app:C}.

\subsection{Supervision and Expansion}\label{b.6-supervision-and-expansion}

The released Training Resource retains task-specific evidence, a claim, a binary label, a contradiction explanation, and source/anchor/pair-triplet linkage metadata. It contains 5,157 Spec-CF pairs, 3,309 Cand-CF pairs, and 2,807 complete triplets. An anchor may have only one branch or no negative. These structures describe the Training Resource and are distinct from the smaller benchmark consistency subsets evaluated in Table~\ref{tab:e10}.

Count and local-property changes supply geometric counterexamples, while marked-region substitutions provide target-binding counterexamples. Affordance and grasp labels add functional constraints. Together, these routes support supervision that connects each judgment to a specific requirement and its evidence.

\section{Metrics}\label{app:C}\label{appendix-c.-metrics}

\subsection{Answer Scores and Aggregation}\label{c.1-answer-scores-and-aggregation}

For the True and False subsets of task \(t\),

\begin{equation}
\mathrm{Acc}_{T,t}=\frac{\sum_{i\in\mathcal D_{T,t}}\mathbf1[\hat y_i=\mathrm{True}]}{|\mathcal D_{T,t}|},\qquad
\mathrm{Acc}_{F,t}=\frac{\sum_{i\in\mathcal D_{F,t}}\mathbf1[\hat y_i=\mathrm{False}]}{|\mathcal D_{F,t}|}.
\end{equation}

Balanced Accuracy averages these two rates. OA averages task-level BA over ten tasks. Human--model differences use the same examples, and are reported in percentage points. Tables display percentage scores, rates, differences, and interval endpoints to one decimal place, while counts remain integers. Aggregates and differences are not recomputed from the rounded table entries.

\subsection{Reference-Semantic Explanation Scores}\label{c.2-reference-semantic-explanation-scores}

Let \(J_i\) indicate explanation acceptance for a negative example. Then

\begin{equation}
\mathrm{FalseJoint}_t=\frac{\sum_{i\in\mathcal D_{F,t}}\mathbf1[\hat y_i=\mathrm{False}]J_i}{|\mathcal D_{F,t}|},
\end{equation}

\begin{equation}
\mathrm{FalseExpAcc}_t=\frac{\sum_{i\in\mathcal D_{F,t}}\mathbf1[\hat y_i=\mathrm{False}]J_i}{\sum_{i\in\mathcal D_{F,t}}\mathbf1[\hat y_i=\mathrm{False}]}.
\end{equation}

FalseJoint measures joint success over all negatives; FalseExpAcc measures acceptance among correct rejections. ExpBalAcc averages True accuracy and FalseJoint. ExpOA averages the ten ExpBalAcc scores. FalseExpAcc is undefined when no negative is correctly rejected; FalseJoint is then zero.

The frozen Judge receives the task, claim, model explanation, and trusted reference explanation. It does not receive model identity or visual evidence. Acceptance requires the candidate explanation to identify the same violated fact and contradiction as the reference. True answers do not undergo an analogous check. Appendix~\ref{d.4-explanation-judge} specifies the Judge implementation.

\subsection{Counterfactual Pairs and Triplets}\label{c.3-counterfactual-pairs-and-triplets}

Here, \(\mathcal P_{b,t}\) is the set of validated pairs for task \(t\) and direction \(b\) (\(S\) for specification-side and \(C\) for candidate-side), while \(\mathcal R_{\triangle,t}\) contains complete validated triplets. Candidate-side pairs require identical model-visible claims; specification-side pairs preserve the candidate evidence. For pair \(p\), \(A(p)\) and \(b(p)\) identify its anchor and negative. For triplet \(r\), \(A(r)\), \(S(r)\), and \(C(r)\) identify the three records.

For example \(i\), let \(q_i=\mathbf1[\hat y_i=y_i]\). Answer-mode correctness is \(u_i^{(0)}=q_i\), while

\begin{equation}
u_i^{(1)}=\begin{cases}q_i,&y_i=\mathrm{True},\\q_iJ_i,&y_i=\mathrm{False}.\end{cases}
\end{equation}

For direction \(b\in\{S,C\}\) and mode \(h\in\{0,1\}\),

\begin{equation}
\mathrm{Pair}_{b,t}^{(h)}=\frac1{|\mathcal P_{b,t}|}\sum_{p\in\mathcal P_{b,t}}u_{A(p)}^{(h)}u_{b(p)}^{(h)},\qquad
\mathrm{Triplet}_{t}^{(h)}=\frac1{|\mathcal R_{\triangle,t}|}\sum_{r\in\mathcal R_{\triangle,t}}\prod_{b\in\{A,S,C\}}u_{b(r)}^{(h)}.
\end{equation}

Matched directional pairs use the same triplets, while all-available pairs use every eligible pair in that direction. Table~\ref{tab:e10} pools 661 specification-side pairs, 727 fixed-claim candidate-side pairs. Triplet consistency is evaluated on 615 complete triplets from six tasks, excluding Grasp Validity, which provides candidate-side pairs but no specification-side pairs or complete triplets.
Fine-grained Geometry contains 30 reference-geometry pairs that are
analyzed separately and are not included in these consistency metrics.
Identity Preservation and Motion Understanding are also excluded from
the pair/triplet consistency analysis.

\subsection{Motion-Issue Accuracy}\label{c.4-motion-issue-accuracy}

Motion Understanding four-class Exact is correct only when the binary answer and primary issue match the reviewed label. Its classes are \emph{Correct}, \emph{Unnatural Motion}, \emph{Incorrect Motion}, and \emph{Deformation Artifacts}. Four-class Macro Recall averages recall across these classes. Joint explanation recall within an issue requires both a correct rejection and an accepted explanation, and differs from binary rejection recall.

\subsection{Uncertainty}\label{c.5-uncertainty}

For the ten-task main results, we use 10,000 bootstrap replicates. Native clusters are resampled independently within each task, retaining linked branches together; task scores are then averaged with equal weight. Fine-grained Geometry positives and negatives are resampled jointly by T-LESS object pair. Shared cluster draws are used across models. Table~\ref{tab:e2} reports marginal intervals, not significance tests for arbitrary model pairs.

For the 400-example human comparison, 20,000 paired replicates resample source clusters within each task and apply the same draws to human and GPT-6 Astra responses. GPT-6 Astra is the fixed comparator, selected by its overall point estimate; the best model is not reselected within replicates. These intervals describe the same-subset comparison and retain the Motion Understanding input-modality difference.

Generation gain intervals resample the forty input pools, retaining all sixteen candidates and final labels within each sampled input and using the same draws for Random and Oracle. The intervals characterize sampling uncertainty, not Judge bias, API drift, or changes in evidence format.

\section{Evaluation and Analysis Interfaces}\label{app:D}\label{appendix-d.-evaluation-and-analysis-interfaces}

\subsection{Subject Models}\label{d.1-subject-models}

The ten subject models use a shared answer-and-explanation contract. Direct inference returns a boolean \texttt{answer} and a concise \texttt{explanation}; Motion Understanding additionally returns \texttt{issue}. All 2,581 examples contribute evaluation records for each model, giving 25,810 model--example records. Empty or invalid answers count as incorrect.

\begin{table}[htbp]
\caption{Evaluated model variants, parameter counts, and Motion Understanding inputs. B denotes billions of parameters; parenthesized active counts retain the A10B/A3B variant information. Dashes indicate that the experiment records do not specify parameter counts.}
\label{tab:d1}
\centering
\begin{adjustbox}{max width=\linewidth}
\begin{tabular}{@{}lrr@{}}
\toprule
Model & Parameters (B) & Motion Understanding input \\
\midrule
GPT-6 Astra & --- & 16-frame sheet \\
GPT-5.6-Sol & --- & 16-frame sheet \\
Gemini-3.8-Flash & --- & Native MP4 \\
Gemma-4-31B & 31 & Native MP4 \\
Qwen3.5-122B-A10B & 122 (10 active) & Native MP4 \\
Qwen3.6-35B-A3B & 35 (3 active) & Native MP4 \\
InternVL3.5-38B & 38 & 16-frame sheet \\
GLM-4.6V-106B & 106 & 16-frame sheet \\
OmniVerifier-7B & 7 & 16-frame sheet \\
Qwen2.5-VL-7B-Instruct & 7 & 16-frame sheet \\
\bottomrule
\end{tabular}
\end{adjustbox}
\end{table}

\begin{table}[htbp]
\caption{Model/service identifiers, precision, and inference concurrency.}
\label{tab:d2}
\centering\setlength{\tabcolsep}{5.5pt}
\begin{adjustbox}{max width=\linewidth}
\begin{tabular}{@{}lrrr@{}}
\toprule
Model & Recorded model / service & Recorded precision & Concurrency \\
\midrule
GPT-6 Astra & gpt-6-astra-2026-09-03 & API service & 8 \\
GPT-5.6-Sol & gpt-5.6-sol & API service & 8 \\
Gemini-3.8-Flash & gemini-3.8-flash & API service & 8 \\
Gemma-4-31B & local-gemma4-31b & NVFP4 & 2 \\
Qwen3.5-122B-A10B & local-sol & FP8 & 8 \\
Qwen3.6-35B-A3B & local-flash & FP8 & 4 \\
InternVL3.5-38B & local-internvl35-38b & BF16 & 2 \\
GLM-4.6V-106B & local-terra & FP8 & 4 \\
OmniVerifier-7B & OmniVerifier-7B & BF16 & 4 \\
Qwen2.5-VL-7B-Instruct & Qwen2.5-VL-7B-Instruct & BF16 & 4 \\
\bottomrule
\end{tabular}
\end{adjustbox}
\end{table}

Tables~\ref{tab:d1} and~\ref{tab:d2} specify the evaluated variants, numerical precision, concurrency, and motion inputs.

Motion Understanding inputs follow each model's supported interface: models with native video support receive MP4 videos, while the others receive ordered sixteen-frame image sheets. This difference reflects input-interface constraints. Nevertheless, the two formats are not equivalent: frame sheets preserve sampled poses and their order but may omit continuous-motion and playback-speed cues. We therefore explicitly report the input format for each model. Human evaluation is conducted with videos provided for Motion Understanding.
Human evaluation uses 40 examples per task. Human--model comparisons use this same subset. Full-benchmark model results use all examples.

\subsection{Fact and Explanation Analysis}\label{d.2-fact-and-explanation-analysis}

Fact-level analysis retains the original claim as input and assigns the decision-relevant property to one of thirteen factual categories. Pair-based comparisons contribute an anchor and a negative decision, whereas Identity and Motion Understanding contribute one decision per native example. The assignment describes factual content rather than an exhaustive decomposition of every claim (Appendix~\ref{app:H}).

Fact groupings describe observable decisions and rejection reasons rather than hidden reasoning processes.

\subsection{Role Separation}\label{d.3-role-separation}

Subject models produce predictions, and the reference-semantic Judge scores correct-negative explanations. Human baseline participants answer without reference labels or model predictions. Separate human review examines target-specific explanation validity, annotates generation candidates, and supplies recursive feedback.

\subsection{Explanation Judge Implementation}\label{d.4-explanation-judge}

The explanation Judge is Qwen3.5-122B-A10B-FP8. It checks whether each correctly rejected explanation identifies the violated fact and contradiction in the reference, using task and claim text but no images or subject-model identifiers. Table~\ref{tab:judge_config} gives its configuration. This checkpoint also appears among the evaluated subject models.

\begin{table}[htbp]
\caption{Explanation-Judge configuration. The Judge checks reference-semantic agreement for correctly rejected negatives.}
\label{tab:judge_config}
\centering
\begin{adjustbox}{max width=\linewidth}
\begin{tabular}{@{}ll@{}}
\toprule
Setting & Value \\
\midrule
Checkpoint / precision & Qwen3.5-122B-A10B / FP8 \\
Serving framework & vLLM \\
Inputs & Task, claim, model explanation, reference explanation \\
Excluded inputs & Visual evidence and subject-model identity \\
Eligible responses & Correctly rejected negatives \\
Temperature / top-$p$ & 0 / 1 \\
Maximum output tokens / seed & 512 / 42 \\
Thinking / concurrency & Disabled / 8 \\
Output format & Strict JSON schema \\
Technical retries / format repairs & At most 5 / at most 3 \\
\bottomrule
\end{tabular}
\end{adjustbox}
\end{table}

\section{Additional Model Results}\label{app:E}\label{appendix-e.-additional-model-results}

\subsection{Overall Decision Behavior}\label{e.1-overall-decision-behavior}

\begin{table}[htbp]
\caption{Overall decision behavior on the ten-task benchmark (\%). OA averages task-level Balanced Accuracy with equal task weights.}
\label{tab:e1}
\centering
\begin{adjustbox}{max width=\linewidth}
\begin{tabular}{@{}lrrrr@{}}
\toprule
Model & Raw accuracy & True recall & False recall & OA \\
\midrule
GPT-6 Astra & 71.5\% & 89.0\% & 61.0\% & 69.4\% \\
GPT-5.6-Sol & 64.9\% & 84.2\% & 53.3\% & 64.9\% \\
Gemini-3.8-Flash & 73.1\% & 72.6\% & 73.4\% & 68.4\% \\
Gemma-4-31B & 62.0\% & 72.3\% & 55.7\% & 62.6\% \\
Qwen3.5-122B-A10B & 59.8\% & 73.7\% & 51.5\% & 61.6\% \\
Qwen3.6-35B-A3B & 52.2\% & 86.4\% & 31.7\% & 59.5\% \\
InternVL3.5-38B & 57.7\% & 60.1\% & 56.2\% & 56.7\% \\
GLM-4.6V-106B & 47.9\% & 89.9\% & 22.6\% & 55.0\% \\
OmniVerifier-7B & 60.7\% & 20.7\% & 84.8\% & 52.0\% \\
Qwen2.5-VL-7B-Instruct & 62.7\% & 2.4\% & 98.9\% & 50.2\% \\
\bottomrule
\end{tabular}
\end{adjustbox}
\end{table}

Table~\ref{tab:e1} reveals substantial differences in decision balance across models. Gemini-3.8-Flash achieves similar True and False recall (72.6\% and 73.4\%), whereas GLM-4.6V-106B favors True decisions and Qwen2.5-VL-7B-Instruct strongly favors False decisions. Manual inspection of the sampled responses of highly imbalanced outputs, including those from Qwen2.5-VL-7B-Instruct, confirmed that the observed skew reflects the models' actual answers rather than errors in our judgment procedure.

\subsection{Explanation-Aware Verification}\label{e.2-explanation-aware-verification}

\begin{table}[htbp]
\caption{Aggregate scores and 95\% cluster-bootstrap intervals on the ten-task benchmark. Scores and interval endpoints are percentages; Drop is in percentage points. FalseExpAcc is explanation acceptance among correct rejections.}
\label{tab:e2}
\centering\small\setlength{\tabcolsep}{4pt}
\begin{adjustbox}{max width=\linewidth}
\begin{tabular}{@{}lrrrrrr@{}}
\toprule
Model & OA & 95\% CI & ExpOA & 95\% CI & Drop (pp) & FalseExpAcc \\
\midrule
GPT-6 Astra & 69.4 & [68.0, 70.8] & 65.8 & [64.4, 67.2] & 3.6 & 87.1 \\
GPT-5.6-Sol & 64.9 & [63.2, 66.6] & 59.6 & [58.0, 61.2] & 5.3 & 82.1 \\
Gemini-3.8-Flash & 68.4 & [66.9, 69.9] & 57.8 & [56.2, 59.4] & 10.6 & 74.2 \\
Gemma-4-31B & 62.6 & [60.9, 64.4] & 52.2 & [50.3, 54.1] & 10.4 & 68.5 \\
Qwen3.5-122B-A10B & 61.6 & [60.2, 63.0] & 48.9 & [47.2, 50.5] & 12.7 & 59.8 \\
Qwen3.6-35B-A3B & 59.5 & [57.9, 61.0] & 51.3 & [49.8, 52.8] & 8.1 & 62.9 \\
InternVL3.5-38B & 56.7 & [55.3, 58.0] & 44.5 & [42.9, 46.1] & 12.2 & 54.0 \\
GLM-4.6V-106B & 55.0 & [53.8, 56.3] & 50.9 & [49.7, 52.1] & 4.1 & 62.6 \\
OmniVerifier-7B & 52.0 & [51.0, 53.1] & 30.0 & [28.6, 31.5] & 22.0 & 48.3 \\
Qwen2.5-VL-7B-Instruct & 50.2 & [49.4, 51.0] & 22.1 & [20.8, 23.5] & 28.1 & 49.0 \\
\bottomrule
\end{tabular}
\end{adjustbox}
\end{table}

\begin{table}[htbp]
\caption{Explanation-aware Balanced Accuracy by task (\%). The two blocks show complementary sets of five models.}
\label{tab:e3}
\centering\footnotesize\setlength{\tabcolsep}{3pt}
\begin{tabularx}{\textwidth}{@{}>{\raggedright\arraybackslash}p{0.23\textwidth}*{5}{>{\centering\arraybackslash}X}@{}}
\toprule
Task & \shortstack{GPT-6\\Astra} & \shortstack{GPT-5.6-\\Sol} & \shortstack{Gemini-3.8-\\Flash} & \shortstack{Gemma-4-\\31B} & \shortstack{Qwen3.5-\\122B-A10B} \\
\midrule
Object & 70.3\% & 70.3\% & 67.4\% & 68.6\% & 63.7\% \\
Part & 67.2\% & 62.3\% & 57.6\% & 50.7\% & 53.4\% \\
Material & 65.0\% & 58.1\% & 60.6\% & 60.0\% & 61.6\% \\
Part Grounding & 86.2\% & 81.1\% & 79.0\% & 64.4\% & 58.3\% \\
Part Relation & 88.0\% & 78.1\% & 84.9\% & 69.8\% & 66.1\% \\
Affordance Region & 73.1\% & 60.3\% & 72.1\% & 48.5\% & 48.4\% \\
Grasp Validity & 51.0\% & 43.5\% & 12.5\% & 38.5\% & 26.5\% \\
Identity Preservation & 55.5\% & 50.5\% & 55.5\% & 56.5\% & 52.5\% \\
Fine-grained Geometry & 50.0\% & 46.7\% & 50.0\% & 33.3\% & 41.7\% \\
Motion Understanding & 51.5\% & 45.5\% & 38.1\% & 32.1\% & 16.4\% \\
\bottomrule
\end{tabularx}
\par\medskip
\begin{tabularx}{\textwidth}{@{}>{\raggedright\arraybackslash}p{0.23\textwidth}*{5}{>{\centering\arraybackslash}X}@{}}
\toprule
Task & \shortstack{Qwen3.6-\\35B-A3B} & \shortstack{InternVL3.5-\\38B} & \shortstack{GLM-4.6V-\\106B} & \shortstack{OmniVerifier-\\7B} & \shortstack{Qwen2.5-VL-\\7B-Instruct} \\
\midrule
Object & 59.6\% & 56.6\% & 59.3\% & 36.3\% & 22.8\% \\
Part & 54.7\% & 42.2\% & 46.8\% & 27.2\% & 22.5\% \\
Material & 56.0\% & 46.5\% & 55.6\% & 28.7\% & 24.5\% \\
Part Grounding & 61.5\% & 38.0\% & 54.3\% & 25.0\% & 29.7\% \\
Part Relation & 63.5\% & 53.8\% & 54.0\% & 44.6\% & 47.7\% \\
Affordance Region & 47.4\% & 40.9\% & 50.0\% & 33.1\% & 24.2\% \\
Grasp Validity & 42.0\% & 45.5\% & 50.0\% & 33.5\% & 13.5\% \\
Identity Preservation & 53.0\% & 49.0\% & 47.0\% & 4.0\% & 14.0\% \\
Fine-grained Geometry & 43.3\% & 50.0\% & 50.0\% & 50.0\% & 5.0\% \\
Motion Understanding & 32.1\% & 22.4\% & 42.5\% & 17.9\% & 17.2\% \\
\bottomrule
\end{tabularx}
\end{table}

The ordering changes after explanation checking. GPT-5.6-Sol has lower OA than Gemini-3.8-Flash but higher ExpOA.

\subsection{Identity Preservation and Motion Understanding}\label{e.3-identity-fine-grained-geometry-motion-understanding}

\begin{table}[htbp]
\caption{Identity Preservation (\%). True and False recall distinguish acceptance of identity-preserving edits from detection of identity changes. Counts refer to correct rejections and reference-Judge failures.}
\label{tab:e4}
\centering\small\setlength{\tabcolsep}{4pt}
\begin{adjustbox}{max width=\linewidth}
\begin{tabular}{@{}lrrrrrr@{}}
\toprule
Model & True recall & False recall & Answer BA & Correct False & Judge fail & ExpBA \\
\midrule
GPT-6 Astra & 100.0\% & 11.0\% & 55.5\% & 11 & 0 & 55.5\% \\
GPT-5.6-Sol & 100.0\% & 1.0\% & 50.5\% & 1 & 0 & 50.5\% \\
Gemini-3.8-Flash & 100.0\% & 14.0\% & 57.0\% & 14 & 3 & 55.5\% \\
Gemma-4-31B & 95.0\% & 22.0\% & 58.5\% & 22 & 4 & 56.5\% \\
Qwen3.5-122B-A10B & 100.0\% & 5.0\% & 52.5\% & 5 & 0 & 52.5\% \\
Qwen3.6-35B-A3B & 91.0\% & 27.0\% & 59.0\% & 27 & 12 & 53.0\% \\
InternVL3.5-38B & 98.0\% & 1.0\% & 49.5\% & 1 & 1 & 49.0\% \\
GLM-4.6V-106B & 85.0\% & 14.0\% & 49.5\% & 14 & 5 & 47.0\% \\
OmniVerifier-7B & 0.0\% & 100.0\% & 50.0\% & 100 & 92 & 4.0\% \\
Qwen2.5-VL-7B-Instruct & 0.0\% & 100.0\% & 50.0\% & 100 & 72 & 14.0\% \\
\bottomrule
\end{tabular}
\end{adjustbox}
\end{table}

\begin{table}[htbp]
\caption{Identity condition-wise ExpBA (\%).
The four conditions stratify examples by expression and texture nuisance settings.}
\label{tab:e5}
\centering\footnotesize\setlength{\tabcolsep}{3pt}
\begin{tabularx}{\textwidth}{@{}>{\raggedright\arraybackslash}p{0.23\textwidth}*{5}{>{\centering\arraybackslash}X}@{}}
\toprule
Condition & \shortstack{GPT-6\\Astra} & \shortstack{GPT-5.6-\\Sol} & \shortstack{Gemini-3.8-\\Flash} & \shortstack{Gemma-4-\\31B} & \shortstack{Qwen3.5-\\122B-A10B} \\
\midrule
Type 1 & 50.0\% & 50.0\% & 52.0\% & 52.0\% & 50.0\% \\
Type 2 & 56.0\% & 50.0\% & 52.0\% & 50.0\% & 54.0\% \\
Type 3 & 50.0\% & 50.0\% & 50.0\% & 52.0\% & 50.0\% \\
Type 4 & 66.0\% & 52.0\% & 68.0\% & 72.0\% & 56.0\% \\
\bottomrule
\end{tabularx}
\par\medskip
\begin{tabularx}{\textwidth}{@{}>{\raggedright\arraybackslash}p{0.23\textwidth}*{5}{>{\centering\arraybackslash}X}@{}}
\toprule
Condition & \shortstack{Qwen3.6-\\35B-A3B} & \shortstack{InternVL3.5-\\38B} & \shortstack{GLM-4.6V-\\106B} & \shortstack{OmniVerifier-\\7B} & \shortstack{Qwen2.5-VL-\\7B-Instruct} \\
\midrule
Type 1 & 52.0\% & 50.0\% & 50.0\% & 2.0\% & 12.0\% \\
Type 2 & 44.0\% & 46.0\% & 34.0\% & 6.0\% & 16.0\% \\
Type 3 & 52.0\% & 50.0\% & 50.0\% & 4.0\% & 14.0\% \\
Type 4 & 64.0\% & 50.0\% & 54.0\% & 4.0\% & 14.0\% \\
\bottomrule
\end{tabularx}
\end{table}

\begin{table}[htbp]
\caption{Joint rejection-and-explanation recall by motion issue (\%).}
\label{tab:e8}
\centering
\setlength{\tabcolsep}{3pt}
\begin{adjustbox}{max width=\linewidth}
\begin{tabular}{@{}lrrr@{}}
\toprule
Model & Unnatural & Incorrect & Deformation \\
\midrule
GPT-6 Astra & 13.6\% & 35.7\% & 0.0\% \\
GPT-5.6-Sol & 4.6\% & 25.0\% & 0.0\% \\
Gemini-3.8-Flash & 9.1\% & 10.7\% & 0.0\% \\
Gemma-4-31B & 31.8\% & 32.1\% & 0.0\% \\
Qwen3.5-122B-A10B & 13.6\% & 35.7\% & 0.0\% \\
Qwen3.6-35B-A3B & 9.1\% & 21.4\% & 0.0\% \\
InternVL3.5-38B & 18.2\% & 64.3\% & 0.0\% \\
GLM-4.6V-106B & 18.2\% & 17.9\% & 0.0\% \\
OmniVerifier-7B & 40.9\% & 28.6\% & 0.0\% \\
Qwen2.5-VL-7B-Instruct & 68.2\% & 28.6\% & 0.0\% \\
\bottomrule
\end{tabular}
\end{adjustbox}
\end{table}

Identity BA ranges from 49.5\% to 59.0\%, and the True/False recalls reveal
strong acceptance biases in several models. Table~\ref{tab:e5}
further reports condition-wise ExpBA under different expression and texture settings. Motion Understanding yields 428 correct rejections and 134 accepted explanations (31.3\% conditional acceptance); its deformation negatives account for 88 correct rejections but no accepted explanation.

\subsection{Native 3D Baselines}\label{e.4-native-3d-baselines}

We additionally evaluate two native point-cloud language models under the
same binary verification interface, as shown in
Table~\ref{tab:native_3d_baselines}.
PointLLM-v1.2-7B~\citep{xu2024pointllm} collapses to all-True predictions,
yielding 50.0\% Balanced Accuracy with zero False recall.
PointLLM-R-7B~\citep{chen2026pointllmr} achieves only 8.3\% Balanced Accuracy,
largely because most responses fail to emit the required binary judgment.
These responses typically terminate normally with EOS but instead produce
empty answer fields or free-form object descriptions. For example, when
asked to verify a claim about a dome-shaped body, a pedestal base, and four
rectangular side openings, the model replies that
``\emph{This object shows a green and black hat.}''
without judging the claim. Its parseable predictions are also strongly
biased toward True. Neither native model succeeds on any eligible
counterfactual pair or triplet.

\begin{table}[htbp]
\caption{Diagnostic results of native point-cloud language models (\%).
Answer BA denotes binary Balanced Accuracy. Invalid or unparseable
binary outputs are counted as incorrect.}
\label{tab:native_3d_baselines}
\centering
\small
\setlength{\tabcolsep}{4pt}
\begin{adjustbox}{max width=\linewidth}
\begin{tabular}{@{}lrrrrrrr@{}}
\toprule
Model & Parse & True Recall & False Recall & Answer BA
& Spec Pair & Cand Pair & Triplet \\
\midrule
PointLLM-v1.2-7B
& 100.0 & 100.0 & 0.0 & 50.0
& 0.0 & 0.0 & 0.0 \\

PointLLM-R-7B
& 16.7 & 16.7 & 0.0 & 8.3
& 0.0 & 0.0 & 0.0 \\
\bottomrule
\end{tabular}
\end{adjustbox}
\end{table}

\subsection{Pair and Triplet Consistency Across Static Tasks}\label{e.5-seven-task-pair-and-triplet-consistency}

The comparison set contains 661 specification-side pairs with fixed
evidence and 727 candidate-side pairs with identical model-visible claims
across seven static tasks. The 615 complete triplets come from six of
these tasks; Grasp Validity contributes candidate-side pairs only and
therefore does not form complete triplets. Fine-grained Geometry is
evaluated separately using 30 reference-geometry pairs, while Identity
Preservation and Motion Understanding do not participate in these
family-consistency metrics.

\begin{table}[htbp]
\caption{Pair and triplet consistency across the static-task subset (\%). Spec Pair uses 661 fixed-evidence pairs, Cand Pair uses 727 fixed-claim pairs, and Triplet uses 615 complete comparisons. Explanation-aware variants additionally require accepted negative explanations. The directional pair columns use all eligible pairs, not a common matched subset. Bold marks the best model score.}
\label{tab:e10}
\centering
\setlength{\tabcolsep}{3pt}
\begin{adjustbox}{max width=\linewidth}
\begin{tabular}{@{}lrrrrrr@{}}
\toprule
Model & Spec Pair & ExpSpec Pair & Cand Pair & ExpCand Pair & Triplet & ExpTriplet \\
\midrule
GPT-6 Astra & \textbf{56.7} & \textbf{54.3} & \textbf{60.8} & \textbf{53.5} & 46.5 & \textbf{40.5} \\
GPT-5.6-Sol & 48.9 & 45.1 & 50.2 & 42.5 & 36.4 & 29.8 \\
Gemini-3.8-Flash & 55.1 & 51.9 & 57.9 & 46.8 & \textbf{49.6} & 40.0 \\
Gemma-4-31B & 39.6 & 33.4 & 40.4 & 30.5 & 26.0 & 18.4 \\
Qwen3.5-122B-A10B & 36.3 & 28.4 & 38.5 & 27.1 & 24.4 & 15.1 \\
Qwen3.6-35B-A3B & 23.1 & 18.6 & 29.2 & 19.8 & 9.8 & 7.3 \\
InternVL3.5-38B & 29.3 & 18.9 & 26.8 & 17.6 & 13.8 & 5.9 \\
GLM-4.6V-106B & 17.4 & 13.0 & 18.3 & 12.9 & 8.0 & 3.9 \\
OmniVerifier-7B & 11.2 & 7.7 & 10.7 & 5.6 & 6.2 & 2.8 \\
Qwen2.5-VL-7B-Instruct & 2.4 & 1.2 & 2.2 & 0.7 & 2.4 & 0.5 \\
\bottomrule
\end{tabular}
\end{adjustbox}
\end{table}

Gemini-3.8-Flash has the highest Triplet success, at 49.6\% (305/615), while GPT-6 Astra reaches 46.5\% (286/615). Requiring both negative explanations to pass changes the ordering: Astra achieves 40.5\% (249/615), compared with Gemini's 40.0\% (246/615). Astra also leads both directional pair metrics. These linked-comparison scores measure consistency beyond individual decisions.

\section{Scope and Release}\label{app:F}\label{appendix-f.-scope-and-release}

The benchmark evaluates task-specific evidence from finite views, marked regions, reference comparisons, and animations. Material labels concern visible appearance, and affordance and grasp labels follow the source task rules. Fine-grained Geometry covers thirty objects; Motion Understanding includes both video and frame-sheet inputs. The text-only explanation Judge measures reference agreement, while targeted human inspection checks two specific defect sets. The recursive study uses human feedback and matches output counts rather than computation or human effort. We focus on evaluating verification performance and introducing a structured training resource to support downstream model optimization, which opens rich avenues for future work to explore verifier training.

The release contains two distinct resources: the 2,581-example, ten-task evaluation benchmark and a 14,126-example, seven-task source-disjoint Training Resource. The Training Resource contains 4,011 source groups, 5,660 True anchors, and 8,466 False examples; benchmark source groups and all descendants are excluded, yielding zero benchmark--training source-group overlap. Released manifests retain source indices and pair/triplet links so that counts, grouping, and linkage structure can be recomputed. Source assets and annotations remain subject to their upstream licenses.

\section{Human Evaluation and Generated-Asset Selection}\label{app:G}\label{appendix-g.-human-evaluation-and-generated-asset-selection}

\subsection{Human Evaluation on the Same Task Subsets}\label{g.1-human-evaluation-on-the-same-task-subsets}

Human participants inspect task evidence without reference answers, explanations, model predictions, or construction labels. Each of ten tasks contributes forty examples, for 400 in total. Identity has nineteen True and twenty-one False examples; the other nine tasks each have twenty of each.

\begin{table}[htbp]
\caption{Human evaluation across all ten tasks (\%). Human BA weights tasks equally using unrounded task scores. It is not the pooled accuracy of 400 responses.}
\label{tab:g1}
\centering\scriptsize\setlength{\tabcolsep}{3pt}
\resizebox{\textwidth}{!}{%
\begin{adjustbox}{max width=\linewidth}
\begin{tabular}{@{}lrrrr@{}}
\toprule
Task & Human N & True / False & Human BA & Human evidence \\
\midrule
Object & 40 & 20 / 20 & 95.0 & Frozen object views \\
Part & 40 & 20 / 20 & 82.5 & Frozen object views \\
Material & 40 & 20 / 20 & 90.0 & Frozen object views \\
Part Grounding & 40 & 20 / 20 & 92.5 & Frozen task sheets \\
Part Relation & 40 & 20 / 20 & 95.0 & Frozen task sheets \\
Affordance Region & 40 & 20 / 20 & 90.0 & Frozen task sheets \\
Grasp Validity & 40 & 20 / 20 & 80.0 & Frozen grasp sheet \\
Identity Preservation & 40 & 19 / 21 & 92.9 & 2-by-8 asset comparison \\
Fine-grained Geometry & 40 & 20 / 20 & 80.0 & 2-by-7 photo/render sheet \\
Motion Understanding & 40 & 20 / 20 & 80.0 & Native MP4 and all captions \\
Human OA & 400 & 199 / 201 & 87.8 & Task-specific evidence above \\
\bottomrule
\end{tabular}
\end{adjustbox}
}
\end{table}

\begin{table}[htbp]
\caption{Human and model Balanced Accuracy on the same 400 examples (\%), with forty per task. OA averages the ten task scores. Humans view native video for Motion Understanding; model inputs follow Table~\ref{tab:d1}. Geometry and Motion abbreviate Fine-grained Geometry and Motion Understanding.}
\label{tab:g2}
\centering\scriptsize\setlength{\tabcolsep}{3pt}
\resizebox{\textwidth}{!}{%
\begin{adjustbox}{max width=\linewidth}
\begin{tabular}{@{}lrrrrrrrrrrr@{}}
\toprule
Evaluator & Material & Object & Part & Afford. & Grasp & Ground. & Relation & Identity & Geometry & Motion & OA \\
\midrule
Human & 90.0\% & 95.0\% & 82.5\% & 90.0\% & 80.0\% & 92.5\% & 95.0\% & 92.9\% & 80.0\% & 80.0\% & 87.8\% \\
GPT-6 Astra & 75.0\% & 72.5\% & 67.5\% & 75.0\% & 57.5\% & 85.0\% & 75.0\% & 54.8\% & 50.0\% & 50.0\% & 66.2\% \\
GPT-5.6-Sol & 75.0\% & 75.0\% & 52.5\% & 67.5\% & 57.5\% & 77.5\% & 67.5\% & 52.4\% & 52.5\% & 55.0\% & 63.2\% \\
Gemini-3.8-Flash & 75.0\% & 82.5\% & 60.0\% & 77.5\% & 60.0\% & 77.5\% & 75.0\% & 57.1\% & 50.0\% & 47.5\% & 66.2\% \\
Gemma-4-31B & 75.0\% & 80.0\% & 45.0\% & 55.0\% & 52.5\% & 70.0\% & 55.0\% & 61.9\% & 47.5\% & 65.0\% & 60.7\% \\
Qwen3.5-122B-A10B & 70.0\% & 62.5\% & 52.5\% & 57.5\% & 50.0\% & 72.5\% & 60.0\% & 57.1\% & 57.5\% & 52.5\% & 59.2\% \\
Qwen3.6-35B-A3B & 62.5\% & 62.5\% & 52.5\% & 52.5\% & 57.5\% & 65.0\% & 65.0\% & 58.8\% & 57.5\% & 67.5\% & 60.1\% \\
InternVL3.5-38B & 65.0\% & 60.0\% & 47.5\% & 52.5\% & 52.5\% & 50.0\% & 52.5\% & 49.8\% & 50.0\% & 57.5\% & 53.7\% \\
GLM-4.6V-106B & 65.0\% & 65.0\% & 47.5\% & 50.0\% & 47.5\% & 55.0\% & 52.5\% & 46.9\% & 50.0\% & 57.5\% & 53.7\% \\
OmniVerifier-7B & 50.0\% & 55.0\% & 40.0\% & 37.5\% & 52.5\% & 52.5\% & 45.0\% & 50.0\% & 52.5\% & 50.0\% & 48.5\% \\
Qwen2.5-VL-7B-Instruct & 47.5\% & 55.0\% & 52.5\% & 50.0\% & 50.0\% & 50.0\% & 50.0\% & 50.0\% & 47.5\% & 50.0\% & 50.2\% \\
\bottomrule
\end{tabular}
\end{adjustbox}
}
\end{table}

The overall comparison uses

\begin{equation}
H=\frac1{10}\sum_{t=1}^{10}\mathrm{BA}_{H}(S_t),\qquad
M_m=\frac1{10}\sum_{t=1}^{10}\mathrm{BA}_{m}(S_t),\qquad
\Delta_{H-\mathrm{Astra}}=H-M_{\mathrm{Astra}}.
\end{equation}

On these same examples, human OA is 87.8\% and GPT-6 Astra reaches 66.2\%, giving a 21.6-point difference. The comparison uses the same subsets and labels, with Astra as the fixed comparator; it does not subtract the full-benchmark score in Table~\ref{tab:main}.

Motion Understanding is the only task with non-identical evidence formats. Humans and Gemini-3.8-Flash, Gemma-4-31B, Qwen3.5-122B-A10B, and Qwen3.6-35B-A3B use native MP4. The other six models receive the frozen sixteen-frame sheet. The ten-task result is therefore a same-sample comparison, not a fully matched-evidence comparison.

Identity human accuracy is 92.5\%, while BA is 92.9\% because the subset is not label-balanced. Table~\ref{tab:g3} compares all models on these same forty examples. Gemma-4-31B has the highest model BA at 61.9\%, a 31.0-point gap from humans; GPT-6 Astra reaches 54.8\%, a 38.1-point gap. Human False recall is 85.7\%, compared with 23.8\% for Gemma-4-31B and 9.5\% for GPT-6 Astra. These results distinguish accepting identity-preserving changes from detecting an identity change.

\begin{table}[htbp]
\caption{Identity Preservation on the matched forty-example human subset (19 True / 21 False). All values except N are percentages. Agreement compares a model's answers with the human responses, whereas BA uses the task reference labels. Values are calculated from the class denominators rather than differences between rounded display values.}
\label{tab:g3}
\centering\scriptsize\setlength{\tabcolsep}{3pt}
\resizebox{\textwidth}{!}{%
\begin{adjustbox}{max width=\linewidth}
\begin{tabular}{@{}lrrrrrr@{}}
\toprule
Judge / model & N & True recall & False recall & Accuracy & BA & Agreement with human \\
\midrule
Human & 40 & 100.0 & 85.7 & 92.5 & 92.9 & 100.0 \\
GPT-6 Astra & 40 & 100.0 & 9.5 & 52.5 & 54.8 & 60.0 \\
Gemini-3.8-Flash & 40 & 100.0 & 14.3 & 55.0 & 57.1 & 62.5 \\
GPT-5.6-Sol & 40 & 100.0 & 4.8 & 50.0 & 52.4 & 57.5 \\
Gemma-4-31B & 40 & 100.0 & 23.8 & 60.0 & 61.9 & 67.5 \\
Qwen3.5-122B-A10B & 40 & 100.0 & 14.3 & 55.0 & 57.1 & 62.5 \\
Qwen3.6-35B-A3B & 40 & 84.2 & 33.3 & 57.5 & 58.8 & 65.0 \\
InternVL3.5-38B & 40 & 94.7 & 4.8 & 47.5 & 49.8 & 55.0 \\
GLM-4.6V-106B & 40 & 84.2 & 9.5 & 45.0 & 46.9 & 52.5 \\
OmniVerifier-7B & 40 & 0.0 & 100.0 & 52.5 & 50.0 & 45.0 \\
Qwen2.5-VL-7B-Instruct & 40 & 0.0 & 100.0 & 52.5 & 50.0 & 45.0 \\
\bottomrule
\end{tabular}
\end{adjustbox}
}
\end{table}

On Fine-grained Geometry, human BA is 80.0\% and the best model BA is 57.5\%, giving a 22.5-point task-level gap. On Motion Understanding, Qwen3.6-35B-A3B reaches the highest model BA of 67.5\% against the human 80.0\%. Human four-class exact accuracy is 70.0\%. These task-wise results use the forty-example subsets in Table~\ref{tab:g2}, not the corresponding full-task sets.

Human OA has a 95\% interval of [84.6, 90.8]\%, and Astra's same-subset OA has an interval of [62.4, 70.0]\%. The paired difference is 21.6 points, with a 95\% interval of [16.8, 26.2] points. These estimates use 20,000 within-task source-cluster resamples with the same draws for human and model responses.

\subsection{Public Image Conditions and Fixed Requirements}\label{g.2-public-image-conditions-and-fixed-requirements}

The generation study uses forty source images from the public image-prompt collection shared by 3DGen-Bench and Hi3DBench. Screening retains complete objects with sufficient visible local facts. Existing Hi3DBench quality labels define four historical strata, with ten sources from each.

Human annotators define and review seven claims for each source: one identity claim and six local requirements covering counts, parts, geometry, relations, region binding, and surface appearance. Source images, claim text, and pass criteria are fixed before generation. This requirement-definition stage is separate from the subsequent human review of generated candidates.

\subsection{Generation and Human Annotation}\label{g.3-generation-and-human-annotation}

Each generator provides eight candidates per source, giving 640 assets in total. Hunyuan3D-2.1 uses the Shape-v2-1 + Paint-v2-1 run. Both generators share forty input images and 280 fixed claims. Candidates are rendered with eight 512-by-512 RGB views.

Multiple human reviewers assess the seven facts, severe defects, and fidelity to the source image. Disagreements are resolved through adjudication before scoring. The final annotation set contains 4,480 fact judgments and 640 labels for each asset-level field. All pooled selection metrics use this final set.

\subsection{Selection Metrics and Pooled Results}\label{g.4-selection-metrics-and-pooled-results}

Let \(a_{ijk}=1\) when candidate \(j\) of input \(i\) satisfies fact \(k\). Let \(d_{ij}=1\) indicate a severe defect and \(r_{ij}=1\) indicate fidelity to the source image. The three scores are

\begin{equation}
s_{ij}=\frac17\sum_{k=1}^{7}a_{ijk},\qquad
f_{ij}=\prod_{k=1}^{7}a_{ijk},\qquad
h_{ij}=f_{ij}(1-d_{ij})r_{ij}.
\end{equation}

Thus Strict Full requires all seven facts, no severe defect, and source fidelity. The deterministic selection order is Strict Full, full factual success, fact satisfaction, source fidelity, absence of severe defects, and finally candidate key. It returns one complete candidate \(j_i^\star\) per input. All reported Oracle criteria are measured on that candidate; facts are never combined across assets.

For \(z\in\{s,f,h\}\) and the sixteen-candidate pool \(\mathcal P_i\),

\begin{equation}
B_z=\frac1{40}\sum_{i=1}^{40}\frac1{16}\sum_{j\in\mathcal P_i}z_{ij},\qquad
O_z=\frac1{40}\sum_{i=1}^{40}z_{i j_i^\star}.
\end{equation}

The oracle uses final human labels and measures a post-hoc selection upper bound, not the performance of a deployed verifier. Random is the uniform expectation within each pool. Gain intervals use paired bootstrap resampling of entire input pools.

\begin{table}[htbp]
\caption{Sixteen-candidate selection after multi-reviewer annotation and adjudication. All rows use the same forty inputs, final labels, and selected candidates. Scores are percentages; gains and their 95\% intervals are percentage points.}
\label{tab:g4}
\centering
\setlength{\tabcolsep}{3pt}
\begin{adjustbox}{max width=\linewidth}
\begin{tabular}{@{}lrrrr@{}}
\toprule
Criterion & Random & Oracle & Gain (pp) & Gain 95\% CI (pp) \\
\midrule
Fact satisfaction & 80.8 & 98.2 & 17.4 & [14.4, 20.5] \\
Full factual success (all seven facts) & 26.7 & 90.0 & 63.3 & [53.0, 73.0] \\
Strict Full & 16.6 & 85.0 & 68.4 & [58.0, 77.8] \\
\bottomrule
\end{tabular}
\end{adjustbox}
\end{table}

The full-success oracle covers 36 of forty inputs, while Strict Full covers 34. The gap between these criteria reflects requirements on asset quality and source fidelity beyond the seven factual judgments.

\subsection{Human-Guided Recursive Refinement}\label{g.5-human-guided-recursive-refinement}

The separate recursive experiment uses eight inputs with an unsuccessful initial Hunyuan3D-2.1 candidate. Parallel evaluation has eight independent candidates per input. Recursive refinement starts from the same first candidate and retains all seven facts. A human verifier identifies unmet requirements and supplies the next edit instruction. The updated 2D condition is realized by Hunyuan3D-2.1 regeneration or VecSet-Edit, with a mask when needed. Verification is repeated on the new asset.

\begin{table}[htbp]
\caption{Recursive and parallel results under the recursive study's annotations. Prefix budgets match evaluated 3D-output counts, not computational or human cost. Additional quality and fidelity rows are computed from the final per-asset flags.}
\label{tab:g5}
\centering
\setlength{\tabcolsep}{3pt}
\begin{adjustbox}{max width=\linewidth}
\begin{tabular}{@{}lrr@{}}
\toprule
Setting / success condition & Success count & Success (\%) \\
\midrule
Parallel single candidate, facts only & 5 / 64 & 7.8 \\
Parallel prefix Oracle@$K_i$, facts only & 1 / 8 & 12.5 \\
Parallel Oracle@8, facts only & 3 / 8 & 37.5 \\
Recursive, all seven facts & 8 / 8 & 100.0 \\
Recursive, facts and no severe defect & 7 / 8 & 87.5 \\
Recursive, facts, no severe defect, and source fidelity & 6 / 8 & 75.0 \\
\bottomrule
\end{tabular}
\end{adjustbox}
\end{table}

\begin{table}[htbp]
\caption{Recursive trajectories and parallel pools. Each trajectory reports satisfied facts after generation and every edit. Quality and fidelity refer to the final asset.}
\label{tab:g6}
\centering\scriptsize\setlength{\tabcolsep}{3pt}
\resizebox{\textwidth}{!}{%
\begin{adjustbox}{max width=\linewidth}
\begin{tabular}{@{}lrrrrr@{}}
\toprule
Input & Parallel fact counts & Recursive trajectory & Edits & Severe defect & Source fidelity \\
\midrule
N05 & 4, 5, 5, 5, 6, 6, 5, 4 & 4 \(\rightarrow\) 7 & 1 & No & No \\
N22 & 6, 6, 5, 6, 6, 3, 7, 6 & 6 \(\rightarrow\) 7 & 1 & No & Yes \\
N27 & 6, 6, 6, 6, 6, 6, 7, 6 & 6 \(\rightarrow\) 7 & 1 & No & Yes \\
F07 & 4, 4, 5, 6, 6, 4, 6, 6 & 4 \(\rightarrow\) 7 & 1 & No & Yes \\
N20 & 6, 7, 5, 7, 7, 6, 6, 6 & 6 \(\rightarrow\) 7 & 1 & No & Yes \\
N21 & 6, 6, 5, 5, 5, 6, 6, 5 & 6 \(\rightarrow\) 6 \(\rightarrow\) 7 & 2 & No & Yes \\
N14 & 3, 4, 4, 4, 4, 3, 3, 4 & 3 \(\rightarrow\) 5 \(\rightarrow\) 2 \(\rightarrow\) 3 \(\rightarrow\) 7 & 4 & Yes & Yes \\
N25 & 2, 2, 2, 2, 2, 2, 2, 5 & 2 \(\rightarrow\) 6 \(\rightarrow\) 7 & 2 & No & Yes \\
\bottomrule
\end{tabular}
\end{adjustbox}
}
\end{table}

\begin{table}[htbp]
\caption{Edit backends for each recursive trajectory. All initial candidates use Hunyuan3D-2.1.}
\label{tab:g7}
\centering
\setlength{\tabcolsep}{3pt}
\begin{adjustbox}{max width=\linewidth}
\begin{tabular}{@{}lr@{}}
\toprule
Input & Edit backend sequence \\
\midrule
N05 & Hunyuan3D-2.1 \\
N22 & VecSet-Edit \\
N27 & VecSet-Edit (texture-only) \\
F07 & Hunyuan3D-2.1 \\
N20 & Hunyuan3D-2.1 \\
N21 & Hunyuan3D-2.1; VecSet-Edit (texture-only) \\
N14 & Hunyuan3D-2.1; VecSet-Edit; Hunyuan3D-2.1; Hunyuan3D-2.1 \\
N25 & Hunyuan3D-2.1; Hunyuan3D-2.1 \\
\bottomrule
\end{tabular}
\end{adjustbox}
\end{table}

The edit counts \((1,1,1,1,1,2,4,2)\) total thirteen. Including the initial assets gives twenty-one evaluated outputs. A corresponding parallel prefix uses \(K_i=1+r_i\) candidates, where \(r_i\) is the observed edit count. This post-hoc prefix comparison matches output counts only.

N21 progresses from 6/7 to 6/7 and then 7/7. Regeneration does not fully resolve the grille requirement; the subsequent texture-focused edit does. N14 follows 3, 5, 2, 3, and 7 satisfied facts, showing non-monotonic refinement, and its final severe-defect flag remains positive. N05 satisfies all facts but retains a negative source-fidelity flag. Hence the final fact-only, quality-gated, and quality-plus-fidelity counts are eight, seven, and six, respectively.

The recursive results demonstrate human-guided refinement. The output-count-matched prefixes use the observed edit counts, rather than a preset compute or human-effort budget.

\section{Fine-Grained Facts and Explanation Reliability}\label{app:H}\label{appendix-h.-fine-grained-facts-and-explanation-reliability}

\subsection{Analysis Procedure}\label{h.1-analysis-procedure}

Motivated by fact-oriented evaluation in DSG and FActScore \citep{cho2024dsg,min2023factscore}, we group existing verification decisions by their decision-relevant factual property. Fine Fact labels organize this analysis; the original claim remains the model input, and no new subject-model queries are created.

The assignment contains 1,778 verification units: 1,444 paired comparisons, 200 Identity comparisons, and 134 Motion Understanding sequences. These yield 3,222 decision incidences per model, with 1,611 True and 1,611 False. Anchors can recur across pairs. Fine-grained Geometry contributes thirty object pairs and sixty decisions. These content-based analysis units differ from distinct benchmark examples and from the eligible comparisons used for consistency scoring.

\subsection{Thirteen Fine Fact Categories}\label{h.2-thirteen-fine-fact-categories}

\begin{table}[htbp]
\caption{Thirteen Fine Fact categories.}
\label{tab:h1}
\centering
\small
\setlength{\tabcolsep}{4pt}
\renewcommand{\arraystretch}{1.08}

\begin{adjustbox}{max width=\linewidth}
\begin{tabular}{@{}p{0.27\linewidth}p{0.66\linewidth}@{}}
\toprule
\textbf{Fine Fact} & \textbf{Meaning} \\
\midrule
identity\_existence
& Object or part identity, category, and existence \\

identity\_preserving
& Preservation of identity across an edit \\

quantity
& Number of objects, parts, or repeated elements \\

shape\_geometry
& Global or local shape, openings, thickness, cross-sections, and curvature \\

spatial\_configuration
& Static position, orientation, arrangement, and relative relation \\

connectivity\_contact
& Attachment, contact, separation, intersection, and topology \\

color
& Visible color and its regional distribution \\
\bottomrule
\end{tabular}
\end{adjustbox}

\vspace{5pt}

\begin{adjustbox}{max width=\linewidth}
\begin{tabular}{@{}p{0.27\linewidth}p{0.66\linewidth}@{}}
\toprule
\textbf{Fine Fact} & \textbf{Meaning} \\
\midrule
texture\_appearance
& Texture, material-like appearance, reflectance, and surface finish \\

region\_binding
& Binding a claim to the correct object, part, or region \\

action\_compatibility
& Compatibility of geometry or contact with a requested action \\

motion\_direction
& Direction or kinematic tendency of a movement \\

grasp\_validity
& Validity of a grasp, its contact, and stability \\

motion\_compliance
& Compliance of the complete motion with the caption set without major motion defects \\
\bottomrule
\end{tabular}
\end{adjustbox}
\end{table}

Shape/geometry includes the Fine-grained Geometry examples. Motion compliance includes Motion Understanding; its three issue labels remain within-task diagnostics. Identity preservation differs from identity/existence: the former compares identity before and after editing, while the latter concerns category or presence in a single verification relation.

\subsection{Fact-Wise Verification Performance}\label{h.3-fact-wise-verification-performance}

\begin{table}[htbp]
\caption{Fact-wise performance across ten models (\%). Units are paired comparisons or individual examples; decisions can reuse anchors across pairs. Motion direction contains four pair units.}
\label{tab:h2}
\centering\scriptsize\setlength{\tabcolsep}{3pt}
\resizebox{\textwidth}{!}{%
\begin{adjustbox}{max width=\linewidth}
\begin{tabular}{@{}lrrrrr@{}}
\toprule
Fine Fact & Units & True decisions & False decisions & Mean BA & Mean ExpBA \\
\midrule
identity\_existence & 59 & 59 & 59 & 65.7\% & 59.4\% \\
identity\_preserving & 200 & 100 & 100 & 53.2\% & 43.8\% \\
quantity & 124 & 124 & 124 & 61.5\% & 47.9\% \\
shape\_geometry & 265 & 265 & 265 & 63.1\% & 51.9\% \\
spatial\_configuration & 94 & 94 & 94 & 65.3\% & 60.4\% \\
connectivity\_contact & 103 & 103 & 103 & 70.8\% & 67.8\% \\
color & 102 & 102 & 102 & 66.4\% & 54.6\% \\
texture\_appearance & 98 & 98 & 98 & 64.3\% & 52.6\% \\
region\_binding & 225 & 225 & 225 & 69.4\% & 58.2\% \\
action\_compatibility & 270 & 270 & 270 & 58.1\% & 50.0\% \\
motion\_direction & 4 & 4 & 4 & 58.8\% & 53.8\% \\
grasp\_validity & 100 & 100 & 100 & 55.8\% & 35.7\% \\
motion\_compliance & 134 & 67 & 67 & 53.5\% & 31.6\% \\
\bottomrule
\end{tabular}
\end{adjustbox}
}
\end{table}

\begin{table}[htbp]
\caption{Model-wise Fine Fact ExpBA (\%). The two blocks show complementary sets of five models.}
\label{tab:h3}
\centering\footnotesize\setlength{\tabcolsep}{3pt}
\begin{tabularx}{\textwidth}{@{}>{\raggedright\arraybackslash}p{0.23\textwidth}r*{5}{>{\centering\arraybackslash}X}@{}}
\toprule
Fine Fact & Units & \shortstack{GPT-6\\Astra} & \shortstack{GPT-5.6-\\Sol} & \shortstack{Gemini-3.8-\\Flash} & \shortstack{Gemma-4-\\31B} & \shortstack{Qwen3.5-\\122B-A10B} \\
\midrule
identity\_existence & 59 & 78.0\% & 64.4\% & 72.0\% & 66.9\% & 65.3\% \\
identity\_preserving & 200 & 55.5\% & 50.5\% & 55.5\% & 56.5\% & 52.5\% \\
quantity & 124 & 67.7\% & 65.3\% & 59.3\% & 53.6\% & 47.2\% \\
shape\_geometry & 265 & 65.5\% & 62.5\% & 61.3\% & 55.3\% & 58.1\% \\
spatial\_configuration & 94 & 83.5\% & 72.3\% & 78.7\% & 64.4\% & 58.5\% \\
connectivity\_contact & 103 & 87.4\% & 82.5\% & 84.0\% & 76.2\% & 76.7\% \\
color & 102 & 74.0\% & 64.2\% & 66.7\% & 65.7\% & 64.2\% \\
texture\_appearance & 98 & 60.7\% & 59.2\% & 60.2\% & 60.2\% & 62.2\% \\
region\_binding & 225 & 86.4\% & 81.3\% & 79.3\% & 65.1\% & 58.9\% \\
action\_compatibility & 270 & 73.1\% & 60.6\% & 72.2\% & 48.9\% & 48.7\% \\
motion\_direction & 4 & 37.5\% & 62.5\% & 50.0\% & 75.0\% & 62.5\% \\
grasp\_validity & 100 & 51.0\% & 43.5\% & 12.5\% & 38.5\% & 26.5\% \\
motion\_compliance & 134 & 51.5\% & 45.5\% & 38.1\% & 32.1\% & 16.4\% \\
\bottomrule
\end{tabularx}
\par\medskip
\begin{tabularx}{\textwidth}{@{}>{\raggedright\arraybackslash}p{0.23\textwidth}r*{5}{>{\centering\arraybackslash}X}@{}}
\toprule
Fine Fact & Units & \shortstack{Qwen3.6-\\35B-A3B} & \shortstack{InternVL3.5-\\38B} & \shortstack{GLM-4.6V-\\106B} & \shortstack{OmniVerifier-\\7B} & \shortstack{Qwen2.5-VL-\\7B-Instruct} \\
\midrule
identity\_existence & 59 & 60.2\% & 54.2\% & 58.5\% & 37.3\% & 37.3\% \\
identity\_preserving & 200 & 53.0\% & 49.0\% & 47.0\% & 4.0\% & 14.0\% \\
quantity & 124 & 54.4\% & 39.1\% & 43.1\% & 25.0\% & 23.8\% \\
shape\_geometry & 265 & 55.8\% & 50.0\% & 54.9\% & 33.8\% & 22.1\% \\
spatial\_configuration & 94 & 57.4\% & 47.3\% & 51.1\% & 48.4\% & 42.0\% \\
connectivity\_contact & 103 & 67.5\% & 64.1\% & 59.7\% & 41.3\% & 38.3\% \\
color & 102 & 53.4\% & 47.5\% & 56.4\% & 30.4\% & 23.5\% \\
texture\_appearance & 98 & 59.7\% & 49.0\% & 57.7\% & 29.1\% & 27.6\% \\
region\_binding & 225 & 62.0\% & 38.4\% & 54.9\% & 25.3\% & 30.0\% \\
action\_compatibility & 270 & 47.8\% & 40.9\% & 50.0\% & 33.1\% & 24.3\% \\
motion\_direction & 4 & 62.5\% & 75.0\% & 50.0\% & 62.5\% & 0.0\% \\
grasp\_validity & 100 & 42.0\% & 45.5\% & 50.0\% & 33.5\% & 13.5\% \\
motion\_compliance & 134 & 32.1\% & 22.4\% & 42.5\% & 17.9\% & 17.2\% \\
\bottomrule
\end{tabularx}
\end{table}

Mean ExpBA is lowest for motion compliance (31.6\%), grasp validity (35.7\%), and identity preservation (43.8\%). Connectivity/contact reaches 67.8\%. Motion direction contains only four pair units, so its rate describes a small diagnostic group rather than a stable category ranking.

\subsection{Observable Errors and Fine Facts}\label{h.4-observable-errors-and-fine-facts}

\begin{table}[htbp]
\caption{Answer errors and reference-Judge failures by factual category, pooled across ten models. Error share is the fraction of all answer errors; Judge failures count reference-mismatched explanations among correct rejections.}
\label{tab:h4}
\centering
\setlength{\tabcolsep}{4pt}
\begin{adjustbox}{max width=\linewidth}
\begin{tabular}{@{}lrrrr@{}}
\toprule
Fine Fact & Units & Answer errors & Error share & Judge fail \\
\midrule
action\_compatibility & 270 & 2260 & 18.7\% & 442 \\
shape\_geometry & 265 & 1958 & 16.2\% & 590 \\
region\_binding & 225 & 1379 & 11.4\% & 503 \\
quantity & 124 & 955 & 7.9\% & 338 \\
identity\_preserving & 200 & 936 & 7.8\% & 189 \\
grasp\_validity & 100 & 884 & 7.3\% & 403 \\
texture\_appearance & 98 & 700 & 5.8\% & 230 \\
color & 102 & 686 & 5.7\% & 240 \\
spatial\_configuration & 94 & 653 & 5.4\% & 92 \\
motion\_compliance & 134 & 623 & 5.2\% & 294 \\
connectivity\_contact & 103 & 601 & 5.0\% & 63 \\
identity\_existence & 59 & 405 & 3.4\% & 74 \\
motion\_direction & 4 & 33 & 0.3\% & 4 \\
\bottomrule
\end{tabular}
\end{adjustbox}
\end{table}

Action compatibility, shape/geometry, and region binding contribute the largest shares of answer errors: 18.7\%, 16.2\%, and 11.4\%, respectively. These volumes depend on both category size and error rate; they are distinct from explanation mismatches among correct rejections.

An incorrect fact/value states the wrong property or state. Incorrect target/binding associates a property with the wrong entity or region. Vague reasons do not localize a checkable contradiction. These distinctions describe observable mistakes in reviewed cases, rather than a separately quantified taxonomy of error mechanisms. Tables~\ref{tab:h4}--\ref{tab:h6} report fact-grouped answer errors and reference-semantic Judge outcomes.

\subsection{Correct Rejections with Reference-Mismatched Explanations}\label{h.5-correct-rejections-with-reference-mismatched-explanations}

\begin{table}[htbp]
\caption{Reference-semantic Judge failures among correct rejections on the ten-task benchmark.}
\label{tab:h5}
\centering
\setlength{\tabcolsep}{3pt}
\begin{adjustbox}{max width=\linewidth}
\begin{tabular}{@{}lrrr@{}}
\toprule
Model & Correct False & Judge failures & Failure rate \\
\midrule
GPT-6 Astra & 982 & 127 & 12.9\% \\
GPT-5.6-Sol & 858 & 154 & 17.9\% \\
Gemini-3.8-Flash & 1182 & 305 & 25.8\% \\
Gemma-4-31B & 898 & 283 & 31.5\% \\
Qwen3.5-122B-A10B & 829 & 333 & 40.2\% \\
Qwen3.6-35B-A3B & 510 & 189 & 37.1\% \\
InternVL3.5-38B & 905 & 416 & 46.0\% \\
GLM-4.6V-106B & 364 & 136 & 37.4\% \\
OmniVerifier-7B & 1366 & 706 & 51.7\% \\
Qwen2.5-VL-7B-Instruct & 1594 & 813 & 51.0\% \\
\bottomrule
\end{tabular}
\end{adjustbox}
\end{table}

\begin{table}[htbp]
\caption{Reference-semantic Judge failures by task, pooled across ten models. Counts refer to model responses, not distinct assets.}
\label{tab:h6}
\centering
\setlength{\tabcolsep}{3pt}
\begin{adjustbox}{max width=\linewidth}
\begin{tabular}{@{}lrrr@{}}
\toprule
Task & Correct False & Judge failures & Failure rate \\
\midrule
Object & 1270 & 432 & 34.0\% \\
Part & 1291 & 544 & 42.1\% \\
Material & 1362 & 528 & 38.8\% \\
Part Grounding & 1634 & 503 & 30.8\% \\
Part Relation & 1370 & 83 & 6.1\% \\
Affordance Region & 1155 & 438 & 37.9\% \\
Grasp Validity & 635 & 403 & 63.5\% \\
Identity Preservation & 295 & 189 & 64.1\% \\
Fine-grained Geometry & 48 & 48 & 100.0\% \\
Motion Understanding & 428 & 294 & 68.7\% \\
\midrule
Total & 9,488 & 3,462 & 36.5\% \\
\bottomrule
\end{tabular}
\end{adjustbox}
\end{table}

Across the ten models, 3,462 of 9,488 correct rejections fail the reference-semantic Judge (36.5\%). Astra's rate is 12.9\%, and GPT-5.6-Sol's is 17.9\%. The task-wise rates reach 100.0\% for Fine-grained Geometry, 68.7\% for Motion Understanding, and 64.1\% for Identity.

\subsection{Representative Failure Cases of GPT-6 Astra}\label{h.6-representative-failure-cases-of-gpt-6-astra}

Figures~\ref{fig:teaser} and~\ref{fig:failures} pair selected model responses with human-reviewed contradictions. These cases illustrate factual-value, target-binding, geometric, and motion errors.

\textbf{Part presence.} Astra describes a rounded, arched suitcase handle even though the reviewed asset lacks a handle. The claimed component is absent rather than merely shaped differently.

\textbf{Region binding.} The highlighted furniture region is the support box beneath the cushion. Astra instead describes the padded seating surface, binding a plausible property to the wrong part.

\textbf{Grasp contacts.} Astra describes opposing contacts, whereas the displayed jaws lie on adjacent sides and the source label marks the grasp as invalid.

\textbf{Reference geometry.} Astra accepts the candidate's main geometry despite its fuller, outward-bowed middle, rounder cross-section, and wider side silhouette. References and rendered views localize the structural mismatch.

\textbf{Layer count.} Astra repeats the claim's six internal cake layers; the reviewed reference identifies seven. Correctly describing the topping does not resolve the count error.

\textbf{Direct contact.} Astra describes the cushion as resting directly against the marked backrest, while the reviewed part pair does not directly touch.

\textbf{Material appearance.} Astra recognizes beige mottling but calls the surface stone-like. The reviewed contradiction is its reflective, metallic appearance.

\textbf{Affordance.} Astra accepts pressing for the highlighted camera dial, although human review confirms that this region does not support pressing.

\textbf{Motion naturalness and deformation.} In Figure~\ref{fig:teaser}, Astra recognizes the rightward collapse but misses the nearly rigid fall and limited leg response. In the sleeping sequence in Figure~\ref{fig:failures}, it accepts the motion as natural despite forelimb--muzzle fusion. Both cases use sixteen-frame inputs.